\documentclass[final]{cvpr}

\usepackage{amssymb}		% to get all AMS symbols
\usepackage{amsmath}
\usepackage{amsthm}
\usepackage{graphicx}		% to insert figures
\usepackage[inline]{enumitem}
\usepackage{booktabs}  % nice looking tables
\usepackage[tight-spacing=true]{siunitx}
\usepackage{multirow}
\usepackage{tabularx}
\usepackage{cancel}
\usepackage{thmtools}
\usepackage{float} % to allow forcing figures HERE
\usepackage{afterpage}
\usepackage{microtype}
\makeatletter

\renewcommand\section{\@startsection {section}{1}{\z@}%
                                   {-1.2ex \@plus -2ex \@minus -.2ex}%
                                   {.5ex \@plus.2ex}%
                                   {\normalfont\large\bfseries}}
\renewcommand\subsection{\@startsection{subsection}{2}{\z@}%
                                   {-.6ex \@plus -2ex \@minus -.2ex}%
                                   {.5ex \@plus.2ex}%
	                           {\normalfont\large\bfseries}}
\renewcommand\subsubsection{\@startsection{subsubsection}{3}{\z@}%
                                     {-.5ex\@plus -.2ex \@minus -.2ex}%
                                     {.2ex \@plus .2ex}%
                                     {\normalfont\large\bfseries}}
\let\paragraph\textbf
\makeatother

\def\tightmath{
\abovedisplayskip=4pt plus 2pt minus 1pt 
\abovedisplayshortskip=2pt plus 1pt minus 1pt 
\belowdisplayskip=4pt plus 2pt minus 1pt 
\belowdisplayshortskip=2pt plus 1pt minus 1pt }
\def\crushmath{
\abovedisplayskip=1pt plus 1pt minus 2pt 
\abovedisplayshortskip=1pt plus 1pt minus 2pt 
\belowdisplayskip=1pt plus 1pt minus 2pt 
\belowdisplayshortskip=1pt plus 1pt minus 2pt }
\tightmath

\usepackage{arydshln}

\newcolumntype{C}{X<\centering}
\newcolumntype{P}[1]{>{\centering\arraybackslash}p{#1}}
\newcolumntype{M}[1]{>{\centering\arraybackslash}m{#1}}

\newcommand\FFIL{FFIL\@\xspace}
\newcommand\AFIL{AFIL\@\xspace}
\newcommand\FFOWL{FFOWL\@\xspace}
\newcommand\AFOWL{AFOWL\@\xspace}

\newcommand\capt[3]{\caption[#2]{\label{#1}\textsc{#2} \small\textit{#3}}}
\newcommand\blfootnote[1]{%
  \begingroup
  \renewcommand\thefootnote{}\footnote{#1}%
  \addtocounter{footnote}{-1}%
  \endgroup
}

\declaretheoremstyle[%
  spaceabove=-1.5ex,%
  spacebelow=6pt,%
  headfont=\normalfont\itshape,%
  postheadspace=1em,%
  qed=\qedsymbol%
]{mystyle}

\setitemize{noitemsep,topsep=0pt,parsep=0pt,partopsep=0pt}
\usepackage[listofformat=parens,justification=centering,subrefformat=subparens]{subfig}
\long\def\commentout#1{}
\usepackage{algpseudocode}
\usepackage{esvect}
\usepackage{xcolor}
\usepackage{times}
\usepackage{epsfig}
\usepackage{graphicx}
\usepackage{amsmath}
\usepackage{amssymb}
\usepackage{bbm}
\usepackage{bbold}
\usepackage[ruled,noend]{algorithm2e}
\usepackage{float}
\usepackage{enumitem}
\usepackage{dirtytalk}

\usepackage{amsmath}
\DeclareMathOperator*{\argmax}{arg\,max}

\usepackage[english]{babel}
\usepackage[pagebackref=true,breaklinks=true,colorlinks,bookmarks=false]{hyperref}

\addto\extrasenglish{

}

\usepackage[noabbrev]{cleveref}      % reference object types automatically
\addto\captionsngerman{
    \crefname{figure}{abb.}{abb.}
}

\def\confYear{CVPR 2021}

\begin{document}
\captionsetup[figure]{labelfont={bf},name={Fig.},labelsep=colon}
\captionsetup[table]{labelfont={bf},name={Tab.},labelsep=colon}

\title{Self-Supervised Features Improve Open-World Learning}

\author{
\begin{tabular*}{0.75\textwidth}{c@{\extracolsep{\fill}}c@{\extracolsep{\fill}}c@{\extracolsep{\fill}}}Akshay Raj Dhamija$^\ddag$ & Touqeer Ahmad$^\ddag$ & Jonathan Schwan \end{tabular*}\\
\begin{tabular*}{0.75\textwidth}{c@{\extracolsep{\fill}}c@{\extracolsep{\fill}}c@{\extracolsep{\fill}}}Mohsen Jafarzadeh & Chunchun Li & Terrance E. Boult \end{tabular*}\\
\normalsize{Vision and Security Technology Lab, University of Colorado at Colorado Springs, Colorado Springs}\\
{\tt\small \{adhamija, touqeer, jschwan2, mjafarzadeh, cli,  tboult\}@vast.uccs.edu}
}

\maketitle
\crushmath
% %%%%%%%%% ABSTRACT
\begin{abstract}
This paper identifies the flaws in existing open-world learning approaches and attempts to provide a complete picture in the form of \textbf{True Open-World Learning}. 
We accomplish this by proposing a comprehensive generalize-able open-world learning protocol capable of evaluating various components of open-world learning in an operational setting. 
We argue that in true open-world learning, the underlying feature representation should be learned in a self-supervised manner. 
Under this self-supervised feature representation, we introduce the problem of detecting unknowns as samples belonging to Out-of-Label space. 
We differentiate between Out-of-Label space detection and the conventional Out-of-Distribution detection depending upon whether the unknowns being detected belong to the native-world (same as feature representation) or a new-world, respectively. 
Our unifying open-world learning framework combines three individual research dimensions, which typically have been explored independently, \ie, Incremental Learning, Out-of-Distribution detection and Open-World Learning. Starting from a self-supervised feature space, an open-world learner has the ability to adapt and specialize its feature space to the classes in each incremental phase and hence perform better without incurring any significant overhead, as demonstrated by our experimental results. 
The incremental learning component of our pipeline provides the new state-of-the-art on established ImageNet-100 protocol. 
We also demonstrate the adaptability of our approach by showing how it can work as a plug-in with any of the self-supervised feature representation methods.

\end{abstract}

\blfootnote{
$^\ddag$Contributed equally. 
}

\blfootnote{\url{https://github.com/Vastlab/SSFiOWL}}
% %%%%%%%%% BODY TEXT
\section{Introduction}
\label{sec:intro}

\begin{figure}
\begin{center}
    \subfloat[Representation overlap in supervised feature space]
    {
    \label{fig:teaser_a}
    \includegraphics[width=.45\columnwidth,]{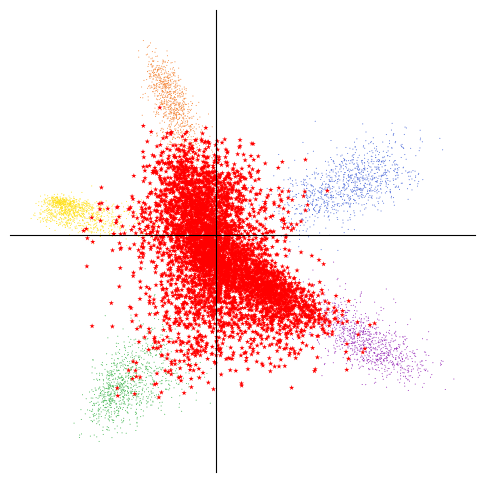}
    }
    \subfloat[Limiting Open-Space risk in self-supervised feature space]{
    \label{fig:teaser_c}\ 
    \includegraphics[width=.42\columnwidth]{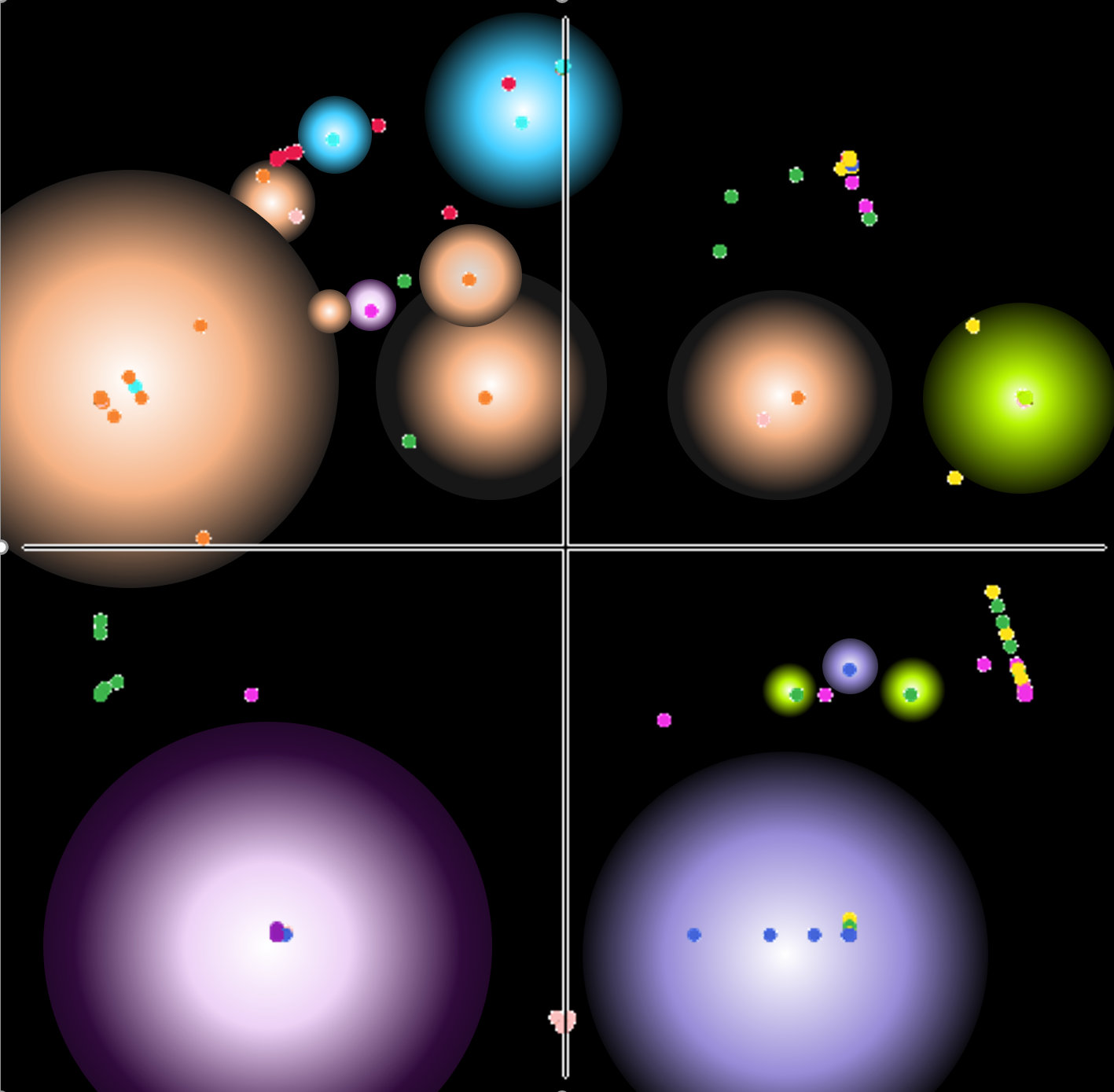}\qquad
    }
\end{center}
\vspace{-.5em}
\capt{fig:teaser}{Why do we recommend self-supervised features for Open-World Learning?}{
\subref*{fig:teaser_a} shows supervised feature space while \subref*{fig:teaser_c} shows probabilities from a self-supervised space (learned by \protect\cite{ji2019invariant}) for the MNIST data.
In \subref*{fig:teaser_a} we show a supervised LeNet++ trained for 5 known classes (0,1,2,3,4) in various colors.
Unknowns (classes 6-9) are overlaid in red, showing the problem of Out-Of-Distribution (OOD) classes overlapping and being difficult to reject as is needed for open-world learning.
Our approach combines self-supervised feature learning with techniques for detecting Out-Of-Label (OOL) space.
The self-supervised feature space has been learned without labels; hence when known classes (0-4) are learned, (6-9) are still out-of-label space but mapped better.
Our approach \subref*{fig:teaser_c} uses distance-based metrics to identify unknown samples as, by default, it considers the region not around any of the known samples as unknown, hence providing a larger black area where unknowns can lie.
}
\end{figure}

Vision systems rarely operate in a closed-world where only the objects seen during training (known objects) are presented to them during the test phase.
%\blfootnote{\url{https://github.com/Vastlab/SSFiOWL}}
Open-world learning consists of classifying the known classes, detecting unknown classes, and incrementally enrolling new classes. A decent attempt towards formalizing open-world learning is made in \cite{bendale2015towards}; however, that work and subsequent paper on the topic \cite{rudd2017extreme} are incomplete in two major aspects. First, instead of addressing true open-world learning, authors \cite{bendale2015towards, rudd2017extreme} end-up performing open-set recognition in conjunction with incremental learning. Specifically, in each incremental phase, instead of learning only with the samples that have been detected as unknowns and accumulated in a memory buffer for an annotator to provide ground-truth labels, they learned with all the samples from newly introduced classes -- more like the conventional incremental learning task. Second, since the feature representation in these approaches is fixed and learned in a supervised manner, they were unable to specialize and adapt the feature space according to the classes in each incremental phase. Herein, we try to bridge these gaps and provide a complete formulation for true open-world learning. Specifically, we propose a comprehensive protocol for true open-world learning where the performance of an open-world agent can be evaluated in each operational phase using the proposed evaluation metrics. Secondly, we argue that the feature space in a true open-world learning scenario should be self-supervised. With each incremental phase, the self-supervised feature space is then adapted according to the classes being introduced in that specific phase.                     

We contend that before operating in an open-world, an agent may gather large amounts of unlabeled data without human intervention that may then be used to learn a feature space via self-supervised training.
This is analogous to how children operate in the world before ever needing to learn ``semantic classes".
Their representation of the world is largely learned with self-supervision.
With labels being provided for some of the objects encountered in the open-world, the system can then learn to identify them without confusing them with other objects for which it had never received the labels due to its robust self-supervised feature representation.
This converts the conventional problem of out-of-distribution detection to a much simpler \textbf{out-of-label space} detection making open-world learning much more effective since the unknowns to be identified are within the distribution of the trained feature space. Fig.\subref{fig:teaser_a} highlights the problem with open-world learning by evolving the supervised feature space from a few known classes. With supervised training for known classes, the resulting feature space overspecializes as noticed by others as well \cite{Han2020Automatically}. This limits the ability to detect out-of-distribution (unknown) classes because of their projection on top of known classes, see \cite{dhamija2018reducing}. This weak feature representation leads to unknowns being classified as knowns. Hence, the underlying open-world system would not attempt to enroll them as new classes.  

We conduct our experiments in two major settings, \ie, incremental learning and open-world learning. Following iCaRL ImageNet-100 protocol \cite{Rebuffi2017}, the first set of our experiments demonstrates that self-supervised features should be an inherent choice for incremental learning. To this end, we first show that by employing only Extreme Value Machines (EVMs) \cite{rudd2017extreme} on top of self-supervised features, we can either perform comparatively or outperform the current state-of-the-art (SOTA) on incremental learning, \ie, PODNet \cite{Douillard2020}. Second, using a two-layer perceptron network, we can adapt the underlying self-supervised features to the specific classes in each incremental phase that results in setting up a \textbf{new SOTA} in incremental learning. Since, the underlying feature representation is pre-learned in a self-supervised manner, incorporating two-layer network does not incur significant additional memory or computational cost. 

We extend our incremental learning frameworks to open-world learning and use thresholding to detect unknowns. We note that any of the existing or emerging unknown detection approaches could be incorporated in our framework but out-of-scope for this paper. Using our proposed open-world protocol and evaluation metrics, we evaluate our open-world agent in both Out-of-Distribution and Out-of-Label space settings. Compared to established incremental learning iCaRL protocol, proposed open-world learning protocol operates in a more operational and practical setting where labels are expected to be available only for a subset of samples for each class. Although demonstrated in a supervised setting, the protocol is extendable and adaptable to semi-supervised and unsupervised variants of open-world learning as well.

\paragraph{Our Contributions}
\begin {enumerate*} [label=\textbf{(\itshape\alph*\upshape)}]
\item Identifying flaws in earlier attempts towards open-world learning and formalizing the problem in its true spirit. 
\item A comprehensive open-world learning protocol and associated evaluation metrics (\autoref{sec:Protocol}).
\item Simplifying unknown detection and incremental learning by evolving underlying feature spaces from self-supervised representation learning. 
\item First to demonstrate that leveraging self-supervised features leads to the new SOTA in incremental learning. 
\item Providing baseline for complete open-world learning in an operational setting using proposed protocol and metrics.  
\end{enumerate*}

\section{Background \& Related Work}
\label{sec:Background}
In an open-world setup, not all classes are available at the initial training phase of a learner; new classes and new instances of old classes are encountered in a temporal manner.
An open-world learner should incrementally learn the new classes while maintaining its performance in classifying the old classes, this area is extensively researched under the name of Incremental Learning.
Moreover, an open-world learner should be capable of not only discriminating between known classes but also identifying and rejecting samples from unknown classes, which are often cast as Out-of-Distribution (OOD) samples and independently studied as OOD detection. 
We discuss both of these problems along with the previous attempts to combine them for open-world learning.

\subsection{Incremental Learning}
\label{subsec:Background_IncrementalLearning}

\paragraph{Fixed Feature Representation}
Incremental learning approaches into this category aim to learn new classes without updating the representation space of new and existing classes.
Nearest Mean Classifier (NMC) \cite{Mensink2013Distance-Based, Mensink2012Metric} is the prime algorithm that represents each class using a prototype vector that is the mean of all the examples seen for that class.
Approaches such as DeeSIL \cite{Belouadah2018DeeSIL} and DeepSLDA \cite{Hayes2020Lifelong} attempt to classify feature representations using independent classifiers such as SVMs.
While one of our incremental learning approaches (\FFIL) still belongs to this small and dying subset of incremental learning methods, we outperform the current supervised approaches by leveraging the better feature representations provided by advances in self-supervised learning.

\paragraph{Adapting Feature Representations}
Most recent incremental learning tries to address \textbf{\textit{catastrophic forgetting}} and \textbf{\textit{concept drift}} by partially/fully re-training the network. 
\cite{Li2018} attempted to address catastrophic forgetting by introducing knowledge distillation in the loss function. 
Another common approach to circumvent catastrophic forgetting is by maintaining \textit{exemplars}, \ie, some samples from old classes are retained. 

The network for the next incremental phase is then trained not only on the new classes but also refined with the exemplars, \eg, \cite{Rebuffi2017, Wu2019, He2020}. 
Generally, the memory budget or the number of exemplars per class is fixed. 
Choosing these exemplars is also an active research area, \eg, methods like \textit{herding} \cite{Welling2009} and \textit{mnemonics} \cite{Liu2020}.
PODNet \cite{Douillard2020}, the \textbf{current SOTA}, studied incremental learning to fight catastrophic forgetting or \textit{rigid-plasticity trade-off} where the network learns to balance between remembering the old classes (\textit{rigidity}) and learning new ones (\textit{plasticity}). 
Unlike other methods \cite{Liu2020,Wu2019,Iscen2020} which typically employ iCaRL protocol \cite{Rebuffi2017} where 5 or more classes are introduced per incremental-task, PODNet \cite{Douillard2020} and \cite{Rebuffi2017,Hou2019,Wu2019} are additionally evaluated on learning one or more classes per task. 
By introducing the accommodation ratio to the cross-distillation loss, He \etal \cite{He2020} tried to address the incremental learning in a more challenging online setting where update time and available data are limited. 
To achieve lifelong learning, each online incremental learning phase was followed by an offline retraining phase where all the data available up to that point was used to retrain the network.
They also maintained an exemplar set and employed herding \cite{Welling2009}. 
Recently, there have been more attempts towards incremental learning \cite{Zhao2020Maintaining, Rajasegaran2020Maintaining} as well as surveying and reviewing the literature on the topic \cite{Luo2020An, Masana2020Class-incremental}. 
In \cite{Belouadah2020A}, a comprehensive evaluation of recently proposed incremental learning approaches \cite{Li2018, Rebuffi2017, Hou2019, Wu2019, Belouadah2019IL2M, Belouadah2020ScaIL, Belouadah2018DeeSIL, Buda2018A, Hayes2020REMIND, Hayes2020Lifelong, mi2020generalized} was conducted, concluding that none of the existing algorithms was significantly better than the others. They also found both memory and incremental step size influenced the relative performance.

\subsection{Out-of-Distribution Detection}
As identified by \cite{Roady2020}, OOD detection approaches can be grouped into 
\begin{enumerate*}[label=\textbf{\alph*)}]
\item inference methods that use an acceptance score function and
\item feature regularization methods that alter the feature representation by changing the network training. 
\end{enumerate*}

Common scoring functions for the first category are based on thresholding the output layers \cite{Hendrycks2017,Li2018,Shu2017}, Helmholtz free energy \cite{liu2020energy}, training one-class networks \cite{Erfani2016,Perera2019} or distance-based scoring \cite{Lee2018,Bendale2016,rudd2017extreme}.
While \cite{Hendrycks2017} directly thresholds SoftMax scores from a network, \cite{Liang2018} incorporates temperature scaling to adjust the SoftMax output.
Lee \etal \cite{Lee2018} employed an acceptance score function by learning a linear classifier to combine the class-conditional Mahalanobis distance metric across multiple CNN layers.
\cite{Erfani2016} and \cite{Perera2019} adapted maximum-margin one-class networks to learn features that enable anomaly detection.
A variant of distance-based classifiers was provided by \cite{Bendale2016,rudd2017extreme} where they used Weibull distributions on the distances between sample features to estimate unknowns.

The second category of OOD detection methods alters the network architecture or training by 
\begin{enumerate*}[label=\textbf{\alph*)}]
\item training one-vs-rest classifiers \cite{Erfani2016,Perera2019},
\item introducing background class regularization \cite{dhamija2018reducing}, or 
\item relying on generative models \cite{Nalisnick2019Do}. 
\end{enumerate*}
Such methods generally need labeled data that do not belong to any of the known classes.
Since labeled data is scarce in the open-world and if data that was earlier learned as none of the known classes needs to be learned as a new class, it will require major network training, which can destroy the online learning concept of open-world learning.

Detecting unknown samples as outliers is an essential component of any open-world framework. 
However, as previously mentioned, an unknown in an open-world setting may be an Out-of-Label space sample or an OOD sample depending upon the underlying feature extractor and the unknown sample. 
We evaluate the unknown detector of our open-world model under both of these distinct problem settings.

\subsection{Open-World Learning}
\paragraph{Novel Class Discovery and Recognition}
Recently \cite{Han2019Learning, Han2020Automatically, Cao2021Open-World} studied some aspects of open-world learning under the title of \textit{novel class discovery and recognition}. 
In \cite{Han2019Learning} authors studied the problem of discovering new classes in unlabeled data via clustering while leveraging the knowledge from labeled data to improve the quality of clustering and to estimate the number of classes in the unlabeled partition. 
This approach operated in an artificial setting where a partition of labeled and unlabeled data sets were pre-assumed and could not isolate labeled samples from unlabeled samples at test time. 
Unlike a true open-world learning setting where new classes are discovered and learned in each incremental phase for several iterations, this approach only discovered and recognized new classes only once and without unknown detection capability. 
Their extended work \cite{Han2020Automatically} is relatively closer to our approach where they also first learned feature representation using both labeled and unlabeled data under a self-supervised auxiliary task, \ie, rotation prediction \cite{gidaris2018unsupervised}. 
A classification head equal to the number of classes in the labeled set and a softmax layer was then added. 
To discriminate between different classes in the unlabeled set, rank statistics was employed to generate pseudo-labels. Subsequently, a different classification head for the unlabeled classes was added to the feature extractor. 
The network was then fine-tuned with these two classification heads by jointly optimizing the two objectives specifically based on labeled and unlabeled partitions. 
It should be noted in both works \cite{Han2019Learning, Han2020Automatically}, the partition of labeled and unlabeled sets is largely unnatural, \eg, in ImageNet experiments, 882 classes belong to the labeled set whereas only 30 classes belong to the unlabeled set. 

The concurrent work presented in \cite{Cao2021Open-World} was based on similar principles as that of \cite{Han2020Automatically} where a feature embedding was learned jointly based on labeled and unlabeled data sets, and linear classification heads were added based on the ground-truth number of classes in the labeled set and the expected number of classes in the unlabeled set. The last layers in the CNN, along with the classification heads, are then finetuned using compound objective based on classification, pairwise similarity, and regularization. 
Differently from \cite{Han2020Automatically}, they operated in a more realistic setting where the number of classes belonging to labeled and unlabeled partitions were more balanced, \eg, in the case of ImageNet, 50 classes belonged to the seen/labeled partition whereas another 50 were considered novel classes in the unlabeled partition. 
Same as \cite{Han2019Learning, Han2020Automatically}, this approach also conducted discovery and recognition of unknown classes just once.              

\paragraph{Open-Set Recognition}
While none of the previous works have focused on true open-world learning, there has been prior work that was very closely related \cite{bendale2015towards,rudd2017extreme}.
These works used open-set recognition in conjunction with incremental learning rather than true open-world learning.
Rather than attempting to find unknowns and then learn them as new classes, they learned new classes in a supervised manner while identifying some of the samples as unknowns.
Furthermore, \cite{rudd2017extreme} used a flawed protocol where the unknown classes to be identified and incrementally learned were a subset of the classes used to train their supervised feature space. The result was inherent feature separation between their known classes and the supposed ``unknown" classes while leading to unrealistically promising good results. 
In one of our open-world approaches (\FFOWL), we extend their framework and bridge the gaps to make it a true open-world approach.   

The issues with prior protocols inspired us to propose a true open-world learning protocol (\autoref{sec:Protocol}) to assess our work and support future research progress.

\section{Our Approach}
\label{sec:ILBP}

In this section, we first discuss the current research trends in incremental learning and then describe two flavors of our proposed incremental approach: (i) \FFIL: employs EVMs directly on the fixed self-supervised features, whereas (ii) \AFIL: incorporates a light-weight two-layer perceptron that adapts the self-supervised features with provided semantic labels at each incremental level. We later demonstrate in our experiments that both versions of our incremental approach outperform the existing SOTA. Next, we extend both our incremental approaches for our fixed feature and adaptive feature open-world learning approaches, \ie, \FFOWL and AFOWL.

\paragraph{Data Availability vs. Label Availability}

In incremental learning at each incremental stage, more labeled data becomes available, and the underlying agent is tasked to learn these new classes while maintaining the learned knowledge for the existing seen classes, \ie, without catastrophic forgetting. In a conventional incremental setting, since input representation is in the form of images, larger networks, \eg, typical to use ResNet-18 for ImageNet-100 protocol \cite{Iscen2020, Liu2020, Douillard2020}, are essential which require longer offline training for each incremental stage. 
Contradictory to that, we start from features trained using self-supervised learning techniques, \ie, instead of using images as input for the subsequent incremental learner, we use self-supervised features. It should be noted that leveraging self-supervised features for downstream tasks such as supervised classification has been demonstrated extensively in recent literature, and in fact, one of the metrics to evaluate the performance of a self-supervised feature extractor is its performance on the downstream task. In a concurrent work \cite{Gallardo2021Self-Supervised}, authors have explored the suitability of self-supervised features for continual learning. However they focused on learning self-supervised feature spaces from data belonging to very small number of classes (10 -- 100), which undermines the purpose of self-supervised learning \ie exploiting abundance of unlabeled data to learn better representations. Herein, we exploit the self-supervised feature spaces for the downstream tasks of incremental and open-world learning where these feature spaces are learned on larger datasets.  
Recently, there have been many attempts towards learning better feature representations under self-supervised learning by defining a \textit{pretext task} \cite{noroozi2016unsupervised, noroozi2018boosting, Zhang2016Colorful, Zhang2017Split-Brain, Larsson2017Colorization, gidaris2018unsupervised, Goyal2019Scaling, Mundhenk2018Improvements, Noroozi2017Representation, Jenni2020Steering, Misra2020Self-Supervised, Jing2020Self-supervised, Gansbeke2020SCAN}, using  \textit{contrastive loss} \cite{He2020Momentum, Chen2020A}, or by \textit{clustering} deep features \cite{Caron2018Deep, Caron2019Unsupervised}.
As demonstrated later in \autoref{sec:experimental_results}, our underlying feature representations can be augmented from any of these advances in the self-supervised learning domain.

\paragraph{Fixed Feature Incremental Learning (FFIL)}
Our FFIL approach belongs to the first category of the incremental learning methods discussed in section \autoref{subsec:Background_IncrementalLearning}, \ie, our feature representation is fixed.
We use self-supervised learning techniques to learn a feature space from unlabeled data that can be used for incrementally learning new classes.
At the zeroth stage of our incremental learning framework, the class labels for initial $C$ classes are provided, and an Extreme Value Machine (EVM) \cite{rudd2017extreme} is fitted for each of these classes. 
The EVM model for each class is comprised of multiple Weibull distributions ($\Phi_i$), where the probability of a sample $x'$ belonging to the distribution may be found with $\Psi\ (x_i,x'; \Phi_i)$ where $x_i$ is the underlying feature vector for $i$-th instance of a class which has been maintained as the representation of that specific class and called an extreme vector.
Please see \cite{rudd2017extreme} for more details.  
At each subsequent incremental step, $K$ new classes are introduced, and new EVM models are fitted for these new classes.
The newly introduced $K$ EVMs are appended to the existing EVM model for $C$ classes, and the process repeats.\par 
As mentioned earlier, recent approaches towards incremental learning are based on retraining the networks with the newly added classes and keeping the exemplars of the old classes to mitigate \textit{catastrophic forgetting}. 
At every incremental stage, multiple rounds of training are essential for good performance of such methods and hence require offline training.
Since our \FFIL classifier is not based on altering the feature representation and does not require back-propagation, we can operate in a true \textbf{online} manner. 
Additionally, \FFIL is not sharing its learning capacity between old and new classes; rather, its recognition capacity is enhanced in the form of new EVMs as new classes are added. It does not suffer from catastrophic forgetting. 

Moreover, the EVM for new classes can be fitted without retaining the exemplars from the old-classes, which corresponds to being a \textbf{zero-exemplar} model. 
We further show that this approach is capable of performing even better when we include exemplars while fitting the new EVMs. 
As the EVM framework is based on retaining the number of extreme vectors which are part of our representation model, the exemplars we use are already part of our model and not explicit images or feature vectors separately maintained in additional memory, so a variant of our approach leveraging the extreme vectors is still a zero-exemplar model. 
It should be noted that earlier methods have a specific memory budget, \eg, 20 exemplars/class to retain the samples of old classes, which is an additional memory requirement than the network weights. 
Approaches like \cite{Iscen2020} try to retain feature vectors instead of images but require an additional adaptation network to project old features to the new feature space, and hence additional training of the secondary network is also required. 
Whereas, our \FFIL approach does not require additional memory to maintain these exemplars as the exemplars (extreme vectors) being used are part of our underlying EVM model.

\paragraph{Adaptive Feature Incremental Learning (AFIL)}
Our AFIL approach belongs to the second and dominant category of the incremental learning approaches discussed in \autoref{subsec:Background_IncrementalLearning}, \ie, our feature representation adapts with each incremental stage. 
However, there is a big difference between our \AFIL approach and other methods, \ie, instead of images, we start from self-supervised feature representations and then adapt these in accordance to semantic labels at each incremental stage. 
Specifically, inspired by classification heads in self-supervised literature, we employ a light-weight two-layer perceptron to adapt the self-supervised features to the classes in each incremental phase.  
At the zeroth stage of our \AFIL approach, the two-layer perceptron is trained with all the available labeled data for initial $C$ classes in the form of feature vectors.  
At each subsequent incremental step, $K$ new classes are introduced, and $K$ new nodes are added in the last layer of our perceptron. To circumvent catastrophic forgetting, like others, we also maintain 20 exemplars per class in the form of self-supervised features and additionally keep the network weights fixed for the nodes for old classes.\par

Unlike FFIL, \AFIL is neither a zero-exemplar nor a truly online approach as it requires exemplars to fight catastrophic forgetting and retraining of the two layer perceptron. However compared to other feature adaptive incremental learning approaches \cite{Douillard2020, Iscen2020, Liu2020} which incorporate larger networks, \eg, ResNet-18/ResNet-32, \AFIL is computational less costly and can be trained much faster while still out-performing its counterparts -- all thanks to evolving from self-supervised feature spaces.

\paragraph{Open World Learning: \FFOWL and \AFOWL }
We extend both of our incremental approaches for open-world learning. We propose \FFOWL and \AFOWL where underlying feature representation is fixed for the former and adaptive for the latter.   
The open-world works in operational phases, each of which is comprised of an uncontrolled real-world environment where an agent finds unknowns and then learns them in an incremental learning stage. 
To this end, we extend \FFIL and \AFIL to detect unknowns by thresholding the Weibull probabilities and SoftMax scores respectively.  

The probability that a query point $x'$ (in the form of feature vector) is associated with a class ${\cal C}_l$ is $\hat P({\cal C}_l|x') = \argmax_{\{i: y_i = {\cal C}_l\}} \Phi\ (x'; \Theta)$. 
Given $\hat P$, we can determine whether $x'$ belongs to an existing known class or is an unknown using:
\begin{equation}
y^* = \begin{cases}\argmax_{1 \le l  \le M} \hat P({\cal C}_l|x^\prime)  %M was undefined , added above

 & \hbox{if } \hat P({\cal C}_l|x^\prime)  \ge \delta \\
\hbox{``unknown"} &  \hbox{Otherwise.}  
\end{cases}
\label{eq:thresholdprob}
\end{equation}

\noindent where $\Phi$ is an EVM model in case of \FFOWL or a two-layer perceptron in case of AFOWL. 
Additionally, the underlying classification model $\Phi$ is parameterized by $\Theta$ which are the Weibull parameters or two layer perceptron parameters for \FFOWL and \AFOWL respectively. 
It should be noted that in FFOWL, $\Theta$ extends with each incremental phase with the detected unknowns that are labeled by the annotator and then enrolled, \ie, new EVMs are appended for newly added classes. 
Whereas for AFOWL, $\Theta$ adapts the underlying feature space according to the newly added classes in each incremental phase.    
After enrolling the newly detected and labeled $k_1$ classes through incremental steps, the \FFOWL/\AFOWL model is now capable of performing multi-class classification for $C+k_1$ classes while detecting the subsequent unknowns. 
It should be noted that a more sophisticated unknown detection approach can be augmented with either of our proposed approaches. However, that dimension is not the focus of the paper and is left for future investigation.

\begin{figure*}[t!]
\begin{center}
\includegraphics[width=\linewidth,trim={0cm 0.9cm 0cm 3cm}]{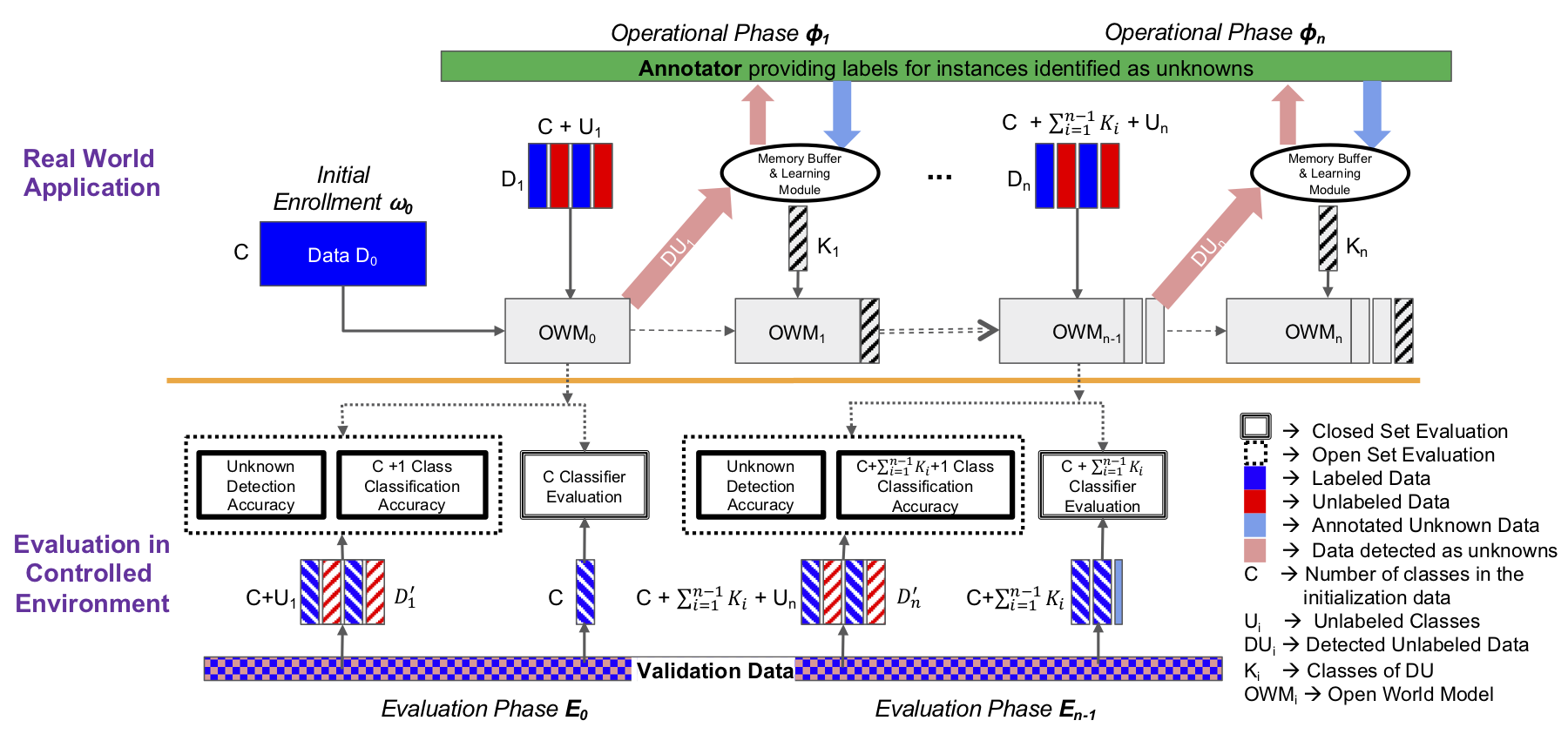}
\end{center}
\capt{fig:fig_protocol}{Open-World Learning}{
PROTOCOL: Initially, the Open-World Model (OWM) is trained with data belonging to $C$ classes. Afterward, in each phase \textbf{$\phi_n$}, the OWM operates in a conventional open-world setting and then learns incrementally. OWM performs multi-class classification on the seen classes in an open-world setting while also detecting and accumulating unlabeled data belonging to new classes in a memory buffer. An annotator then provides the $K_n$ class labels for the detected unlabeled samples (${DU}_n$) present in the memory buffer. The OWM model is then updated for these new $K_n$ classes and is now capable of multi-class classification for $C+\sum_{i=1}^{n}{K_i}$ classes. In each phase, using the validation set, we evaluate our OWM model first for open-world performance, i.e., multi-class classification and unknown detection, and then on the incremental learning task.}
\end{figure*}

\section{Open-World Learning Protocol}
\label{sec:Protocol}
As discussed in \autoref{sec:Background}, open-world learning can be broadly categorized into two sub-domains, out-of-distribution detection and incremental learning.
 
In this section, we bridge the gap between the protocols followed for these two independent research domains by proposing our open-world learning protocol (\autoref{fig:fig_protocol}) that can help advance research in this area of \emph{prime practical importance}.
Any approach addressing the open-world problem has to identify unknown samples, get them labeled by an annotator, and learn them for future encounters.
The most important distinction between an incremental learning problem and an open-world learning problem lies in temporal encounters with unknown data. Compared to out-of-distribution research, the major difference is the continuous change in the data labels between each phase from unknowns to knowns.
To mimic this real-world temporal chain of events, the protocol presents the data to the algorithm in a set of phases which can be broadly classified into three phases for easier understanding.

\paragraph{Initialization Phase \textbf{$\omega_0$}}
Before entering the real-world operational phases, the algorithm needs to go through an initialization phase ($\omega_0$).
In this phase, the protocol provides labeled data $D_0$ from which the algorithm learns to classify $C$ classes.
The resulting open-world model, identified as ${OWM}_0$, is now ready to be deployed in the real-world scenario.

\paragraph{Operational Phase \textbf{$\phi_n$}}
This phase mimics the real-world operation where an agent is responsible for classifying the data into learned classes as well as identifying data belonging to unknown classes.
Here the algorithm operates in phases with some feedback from the human annotator that is simulated in this protocol.
In the operating phase, $\phi_1$, the protocol presents the algorithm with samples not just from the $C$ classes (learned in $\omega_0$) but also from unlabeled/unknown classes $U_1$.
At this point, the model ${OWM}_0$ is responsible for identifying unknown samples from $D_1$ as ${DU}_1$ (detected unknowns) while also correctly classifying any of the samples in $D_1$ that belong to $C$.
The detected unknowns ${DU}_1$ then, depending on the algorithm, may be held in a memory buffer before being sent to an annotator for obtaining labels.
Once the annotator provides labels for all samples in ${DU}_1$, the algorithm now has labeled samples from $K_1$ classes, where $K_1 \subset U_1$.
Using these samples, the algorithm is able to adapt itself to classify classes $C+K_1$.
This protocol continues for various operational phases $\phi_n$, where in each step, the classes in $D_n$ increase as $C+\sum_{i=1}^{n-1} K_i+U_n$.

\paragraph{Evaluation Phase \textbf{$E_n$}}
This phase is aimed at evaluating various aspects of the agent's performance.
We use three major metrics to report a model's performance under the open-world and the closed-world setup.
\begin{enumerate*}[label=\textbf{(\itshape\alph*\upshape)}]
    \item \textit{\textbf{Closed-world Classification Accuracy (CwCA)}}
    The closed-world classification accuracy for any given evaluation phase $E_n$ is measured as the accuracy in terms of the samples belonging to the classes $C+\sum_{i=1}^{n-1}K_i$.
    \item \textit{\textbf{Open-world Performance}}
    In order to measure a model's open-world performance at phase $E_{n-1}$, we feed the model samples from a validation set $D_{n}^{'}$ that contains samples from $C+\sum_{i=1}^{n-1} K_i+U_n$.
    We break down the open-world performance of model $OWM_{n-1}$ in the following two parts. 
    \begin{enumerate*}[label=\textbf{(\itshape\roman*\upshape)}]
    \item \textit{Unknown Detection Accuracy (UDA)}
    This measure provides us an insight towards the algorithm's ability to detect unknowns. 
    It is simply the binary class accuracy for identifying the samples as unknowns.
    \item \textit{Open-world Classification Accuracy (OwCA)}
    The open-world classification accuracy combines the CwCA and UDA by considering the problem to be classifying the sample into Knowns$+1$ classes.
    \end{enumerate*}
\end{enumerate*}

\paragraph{Protocol Variants}
Based on the above descriptions, we propose Open-World-100 (OW-100) and Open-World-500 (OW-500) protocols.
These protocols use images from ImageNet 2012 \cite{Russakovsky2015ImageNet}.
OW-100 protocol uses the same classes as in ImageNet-100 incremental protocol \cite{Rebuffi2017}, while OW-500 keeps extending OW-100 to 500 classes with more batches.
Since the open-world protocols use unknowns to test a model's performance, the algorithms have to follow a basic constraint regarding the training data that is also commonly followed in the out-of-distribution detection research.
The model's underlying feature representations should not be explicitly trained to classify any of the classes in ImageNet 2012 which may be used as unknowns in the protocol.

\paragraph{Protocol Extension and Adaptability}
Although described in a supervised setting, the proposed protocol is applicable and extendable to semi-supervised and unsupervised open-world settings as well. For example, instead of getting annotations from a human annotator for all the discovered unknowns, a subset of unknown samples can be annotated. The detected unknowns can be first clustered and then only representative samples per cluster, \eg, cluster centroids, are annotated by the human operator. Similarly, a human-in-the-loop setting can be avoided altogether and instead, pseudo-labels can be generated by an autonomous annotator, \ie, a clustering method. To this end, such a formulation can benefit further from approaches that estimate the number of clusters or improve clustering quality by leveraging knowledge from labeled/known data, \eg, \cite{Han2019Learning}. We intend to explore these dimensions of open-world learning and variants of our protocol in future work.

\begin{table}[t!]
\capt{tab:quantitative_imagenet}{Average incremental Top-1 accuracy, Ours vs state-of-the-art:}{
Even while using \textbf{zero-exemplars}, \FFIL outperforms PODNet \cite{Douillard2020} (SOTA) on 5 \& 10 class increments.
Additionally, when using exemplars equivalent to all other approaches, \FFIL either out performs \cite{Douillard2020} or comparable across all incremental steps. Further, our \AFIL approach outperforms \cite{Douillard2020} across all the incremental steps by a large margin.
Results with * are from \cite{Hou2019} and with $^\dagger$ generated by \cite{Douillard2020}. 
Results with $^\ddag$ are from \cite{Liu2020} where the original approaches were combined with the \textit{mnemonics} framework. 
}
\centering
\small
\setlength{\arrayrulewidth}{.1em}
\setlength\tabcolsep{3.6pt}
\begin{tabularx}{.9\linewidth}{p{2.7cm}|cccc}
 & 50 {\scriptsize{steps}} & 25 {\scriptsize{steps}} & 10 {\scriptsize{steps}} & 5 {\scriptsize{steps}}\\
 \textbf{New classes per step} & \textbf{1} & \textbf{2} & \textbf{5} & \textbf{10}\\\hline
%  iCaRL* \cite{Rebuffi2017}  & ---   & ---   & 59.53  & 65.04\\
 iCaRL$^\dagger$ \cite{Rebuffi2017}         & 54.97 & 54.56 & 60.90  & 65.56\\
 iCaRL$^\ddag$ \cite{Rebuffi2017}         & --- & 67.12 & 70.50  & 72.34\\ 
 BiC$^\dagger$ \cite{Wu2019} & 46.49 & 59.65 & 65.14  & 68.97\\
 BiC$^\ddag$ \cite{Wu2019} & --- & 69.22 & 70.73  & 71.92\\
%  UCIR\,{\scriptsize (CNN)}* \cite{Hou2019}    & ---   & ---   & 68.09  & 70.47\\
 UCIR\,{\scriptsize (CNN)}$^\dagger$ \cite{Hou2019}     & 57.25 & 62.94 & 67.82  & 71.04\\
 UCIR\,{\scriptsize (CNN)}$^\ddag$ \cite{Hou2019}     & --- & \textbf{69.74} & 71.37  & 72.58\\
 PODNet\,{\scriptsize (CNN)}$^\dagger$ \cite{Douillard2020}  & 62.48 & 68.31 & 74.33 & 75.54\\\hline 
 \textbf{\FFIL}  & \textbf{64.50} & 68.28 & \textbf{75.59} & \textbf{78.09}\\ 
 \textbf{\FFIL}{\scriptsize (\bf Zero Exemplar)} & --- & 61.37 & \textbf{74.43} & \textbf{77.72}\\
 \textbf{\AFIL} &  \textbf{84.86} & \textbf{84.55} & \textbf{84.30} & \textbf{84.28}\\ \hline 
\end{tabularx}
\end{table}

%\section{Experimental Results}
\section{Experimental Results}
\label{sec:experimental_results}
Next we list the details of our experimental setup and results for our incremental and open-world approaches. 

\paragraph{Network Backbone and Datasets}
Unless otherwise stated, all our experiments involve training of underlying self-supervised methods with ResNet-50.
Based on the unlabeled data that was used for training the self-supervised learning approach we make the distinction of an Out-of-Label (OOL) space experiment versus an Out-of-Distribution (OOD) experiment.
For our \textbf{OOL} experiments we trained the self-supervised methods with unlabeled ImageNet 2012 data.
For the \textbf{OOD} experiments we further provide two variants:
\begin{enumerate*}[label=\textbf{\alph*)}]
\item Places2 \cite{zhou2018places} as the training dataset,
\item OpenImages-v6 \cite{Kuznetsova2020The} as the training dataset. 
\end{enumerate*}
Wherever OOD/OOL comparison is applicable, we used \cite{Chen2020Improved} to learn self-supervised features. For backbone variation, we also trained ResNet-18 in same self-supervised manner as that of ResNet-50.

\paragraph{Incremental Learning}
Our comparative results for class incremental learning module are documented in \autoref{tab:quantitative_imagenet} where we compare our approaches against current SOTA in incremental learning \cite{Rebuffi2017,Castro2018,Wu2019,Douillard2020,Liu2020}. 
We follow iCaRL \cite{Rebuffi2017} ImageNet-100 protocol which starts with initially learning 50 classes and then adds classes incrementally. 
While the original protocol added 10 and 5 classes per incremental step, \cite{Douillard2020} explored smaller incremental steps. 
Our experiments with \FFIL/\AFIL utilize features from SwAV \cite{Caron2020Unsupervised} and trained with unlabeled ImageNet data using ResNet-50.
Consistent with the literature, we report the average incremental Top-1 accuracy. 
We compare our approaches to the most recent work \cite{Douillard2020} on incremental learning and additionally from the \textit{mnemonics} approach \cite{Liu2020}.
The numbers reported for each of the earlier approaches are based on 20 exemplars per class.\par 
For our \FFIL approach, we report numbers for both cases, \ie, when the EVM models for new classes are added without information from the old class (zero-exemplar) and when a chosen number of extreme vectors per class (20) serve as the negative examples for the new classes. 
As is apparent from \autoref{tab:quantitative_imagenet}, even with our \textbf{zero-exemplar} model, we are capable of outperforming the latest SOTA \cite{Douillard2020} on 10 \& 5 class increments.
Additionally, when the new EVMs \cite{rudd2017extreme} are fitted using 20 extreme vectors from each of the old classes, we further outperform \cite{Douillard2020} on 1 class incremental step and comparable to it on 2 class increments. 
Our adaptive feature approach (\AFIL) outperforms the existing SOTA \cite{Douillard2020} and understandably our \FFIL approach as well. Consistent with earlier methods, we also maintain 20 exemplars per class in the form of self-supervised feature vectors from old classes.\par 

As discussed in \autoref{sec:ILBP}, both our incremental and open-world approaches are orthogonal to the current advances in self-supervised learning and can benefit from any future advances in this domain. 
To this end, \autoref{tab:diff_feat_incremental1} \& \autoref{tab:diff_feat_incremental2} in the supplemental material demonstrate that both \FFIL and \AFIL can be augmented with any of the recent self-supervised approaches. Further, results for \AFIL in OOL and OOD settings, and with architecture variation can also be found respectively in \autoref{tab:incremental_OOL_OOD} and \autoref{tab:incremental_ResNet50_ResNet18} in the supplemental.\par

\begin{table}[t!]
\capt{tab:openworld_base_results}{Open-World Performance for \FFOWL and \AFOWL}{
Below we report the performance of our \FFOWL \& \AFOWL approaches on the proposed OW-100 and OW-500 protocols.
Based on the unlabeled data used during training of these networks, we report Out-Of-Label (OOL) space and Out-Of-Distribution (OOD) performance.
We report OwCA and CwCA for a fixed UDA at 50\%. 
The reported numbers are average accuracies across all batches. More results including OwCA/CwCA at various fixed UDAs can be found in supplemental.
}
% {\vrule width 0.05em}
\centering
\small
\setlength{\arrayrulewidth}{.1em}
\setlength\tabcolsep{2pt}
\begin{tabularx}{\linewidth}{M{1.8cm}|M{1.cm}|cc|cc}
 \multirow{4}{*}{\textbf{\shortstack[lb]{Protocol\\Details}}} & \multirow{4}{.5cm}{\rotatebox[origin=b]{270}{\textbf{Approach}}} & \multicolumn{4}{c}{\textbf{New classes per step $|U_n|$}}\\\cline{3-6}
 & & \multicolumn{2}{c|}{5} & \multicolumn{2}{c}{10}\\
 & & \multicolumn{2}{c|}{\textit{\scriptsize{\# Knowns/ \# Unknowns}}} & \multicolumn{2}{c}{\textit{\scriptsize{\# Knowns/ \# Unknowns}}}\\
 & & \multicolumn{2}{c|}{\textit{2500/1250}} & \multicolumn{2}{c}{\textit{2500/2500}}\\\cline{3-6}
 & &  OwCA & CwCA & OwCA & CwCA\\\cline{1-6}

 \multirow{2}{*}{\shortstack[lb]{\scriptsize \textbf{OW-100 (OOL)}}}
 & \scriptsize\FFOWL & 65.13 & 69.84 & 60.83 & 69.94 \\  
 & \scriptsize\AFOWL & \textbf{68.32} & \textbf{73.24} & \textbf{64.27} & \textbf{73.84} \\ \cline{1-6} 
 \multirow{2}{*}{\shortstack[lb]{\scriptsize \textbf{OW-100 (OOD)}\\ \scriptsize OpenImages}}  
 & \scriptsize\FFOWL & 41.39 & 44.42 & 40.38 & 46.52 \\ 
 & \scriptsize\AFOWL & \textbf{52.66} & \textbf{56.50} & \textbf{52.19} & \textbf{60.04} \\ \cline{1-6} 
 \multirow{2}{*}{\shortstack[lb]{\scriptsize \textbf{OW-100 (OOD)}\\ \scriptsize Places}}
 & \scriptsize\FFOWL & 34.07 & 36.57 & 33.45 & 38.59 \\ 
 & \scriptsize\AFOWL & \textbf{50.98} & \textbf{54.71} & \textbf{49.33} & \textbf{56.77}\\ \cline{1-6} 
 \multirow{2}{*}{\shortstack[lb]{\scriptsize \textbf{OW-500 (OOL)}}}
 & \scriptsize\FFOWL & 51.63 & 53.11 & 49.45 & 52.31 \\ 
 & \scriptsize\AFOWL & \textbf{54.40} & \textbf{55.95} & \textbf{53.40} & \textbf{56.46}\\ \cline{1-6}  
 \multirow{2}{*}{\shortstack[lb]{\scriptsize \textbf{OW-500 (OOD)}\\ \scriptsize OpenImages}}  
 & \scriptsize\FFOWL & 28.68 & 29.54 & 27.93 & 29.64 \\
 & \scriptsize\AFOWL & \textbf{36.23} & \textbf{37.33} & \textbf{36.24} & \textbf{38.48} \\ \cline{1-6} 
 \multirow{2}{*}{\shortstack[lb]{\scriptsize \textbf{OW-500 (OOD)}\\ \scriptsize Places}}      
 & \scriptsize\FFOWL & 24.41 & 25.13 & 23.67 & 25.10 \\ 
 & \scriptsize\AFOWL & \textbf{32.73} & \textbf{33.75} & \textbf{33.19} & \textbf{35.26} \\\cline{1-6} 
\end{tabularx}
\end{table}

\paragraph{Open-World Learning}

For our open-world learning experiments we use the OpenWorld-100 (OW-100) and OpenWorld-500 (OW-500) protocols proposed in \autoref{sec:Protocol} and summarize the results in \autoref{tab:openworld_base_results} for both variants of OWL.
Since \FFOWL and \AFOWL incorporate different unknown detection methods; we cannot compare them at a fixed threshold. 
Instead we report the OwCA and CwCA  at a fixed UDA percentage that provides a direct comparison between \FFOWL and \AFOWL without requiring the knowledge of specific thresholds for both methods. 
In \autoref{tab:openworld_base_results}, we report OwCA/CwCA at a fixed UDA of 50\%, more results with other UDA percentages can be found in supplemental material. 
From \autoref{tab:openworld_base_results} we observe that it definitely helps improve the classification accuracy if the model has been initialized with data that is closer to the testing world.
For example, OpenImages data (object-centric) is closer to the ImageNet data (object-centric) as compared to the Places2 data (scene-centric) which has less resemblance to ImageNet.
In accordance to the results on incremental task, \AFOWL is consistently outperforming \FFOWL in all settings.
Understandably the performance for both methods decline for OW-500 compared to OW-100 protocol, \ie, increased \textit{catastrophic forgetting} due to a larger number of increments.

\section{Discussion}

Open-world learning strongly depends on the ability to both detect unknown samples and incrementally learn from them once detected. 
This paper shows that using self-supervised features improves the performance of both tasks. 
Our experiments with Fixed Feature Incremental Learning (\FFIL) showed that using appropriate self-supervised feature spaces improves incremental learning under the standard ImageNet-100 protocol. SOTA incremental techniques adapt their fully-supervised representation and reason about what exemplars to keep, yet self-supervised features allowed our \FFIL with zero-exemplars to outperform all of them -- see supplemental material for more examples and discussion. 
Using self-supervised features supported using a deeper network producing better features in turn allowing our EVM-based approach to generally outperform the SOTA when 20 exemplars per class are used. 
On the other hand, adapting the self-supervised features according to the semantic labels at each incremental stage, our \AFIL approach outperforms the existing approaches by a large margin and sets a new \textbf{SOTA}.  
Our incremental approaches satisfy the desired properties defined in \cite{Belouadah2020A}.

Our paper formalizes the novel concept of Out-of-Label (OOL) space detection as a product of evolving open-world learning from self-supervised feature spaces. 
Our \AFOWL experiments provide new SOTA results for open-world learning in both OOL and OOD settings and studied the impact of different self-supervised feature spaces, including ImageNet space, moderately related OpenImages-v6 \cite{Kuznetsova2020The}, and a less related Places2 \cite{zhou2018places} space.
These experiments validate our hypothesis that OOL is easier than OOD.

\blfootnote{
\noindent\textit{\small{\paragraph{Acknowledgement}
This research was  sponsored  by the Defense Advanced Research Projects Agency (DARPA) under HR001120C0055. 
The views contained in this document are those of the authors and should not be interpreted as representing the official policies, either expressed or implied, of the DARPA or the U.S. Government.
}}}

While the proposed approach has advanced the SOTA, there are still some important limitations. 
The performance can be improved by exploring various aspects. 
These include bigger/better self-supervised networks, intelligent selection of exemplars, and using Weibull probabilities during class updates for the \FFIL \& \FFOWL approaches.

\section{Conclusion}
Open-world learning is a growing research sub-field leveraging feature learning, incremental learning, and unknown detection. 
This paper identifies the flaws in earlier attempts towards open-world learning and tries to bridge these gaps by motivating that self-supervised features should be the inherent choice for both incremental and open-world learning. 
Stemming from self-supervised feature spaces, we identify the novel problem of Out-of-Label (OOL) space detection and discriminate it from the conventional Out-of-Distribution (OOD) detection. 
We introduce the first true open-world protocol and its evaluation paradigm, which builds on top of a widely used incremental learning protocol.
Through our experiments and results, we demonstrate that deriving from self-supervised feature spaces, our  fixed feature incremental approach either outperforms or is comparable to the existing SOTA. We further demonstrate that by adapting self-supervised feature with each incremental step using a light weight perceptron, we are able to set a new SOTA in incremental learning outperforming existing SOTA by a large margin.  
Extending our \AFIL approach for open-world learning, we provide the first baseline for true open-world learning in both OOL and OOD settings employing proposed protocol and evaluation metrics.

% \newpage
{\small
\bibliographystyle{ieee_fullname}
\bibliography{nowl}
}

\clearpage
\section*{Supplemental Material}
Here we include the supplementary results and discussion,  first for incremental learning approaches and then for our open-world learning approaches. 

\subsection*{Incremental Learning}
\autoref{tab:diff_feat_incremental1} \& \autoref{tab:diff_feat_incremental2} respectively provide the comparison of using different self-supervised features with \FFIL and \AFIL. In \FFIL approach, the EVM performance is based on parameters \eg distance multiplier (dm) and cover threshold (ct), please see \cite{rudd2017extreme} for more details. For each of the self-supervised feature spaces, we have conducted a grid-search on these parameters on a part of the training set to find the best parameter combinations. The best-found parameter combination for each feature space and each incremental setting is provided as (dm,ct) underneath each average incremental number in \autoref{tab:diff_feat_incremental1}. For the best-performing feature space \ie SwAV, we also provide the parameter variation (\autoref{tab:diff_feat_incremental1} last four rows after the dashed line) to demonstrate the performance sensitivity.      

Since SwAV \cite{Caron2020Unsupervised} is the best-performing feature space for both \FFIL and \AFIL, we also provide the upper bound for SwAV (\autoref{tab:diff_feat_incremental2} last row) wherein each subsequent incremental step, all the data from old classes are also used instead of just using the 20 exemplars per old class. In \autoref{tab:diff_feat_incremental2} (first row) we also provide the results when two-layer perceptron is trained on the feature space learned in a supervised manner only on the data available for batch zero of the incremental learning. This mimics the real-world setting where labelled data is available for limited number of classes and more classes are added incrementally. However adapting larger networks \eg ResNet-18/ResNet-50 with incrementally introduced classes would require offline and longer rounds of training. Others \cite{Iscen2020} have also addressed this by adapting features between incremental rounds.        

\autoref{fig:fig_AFIL_vs_FFIL} provides the visual comparison of the performance of \FFIL and \AFIL for all feature spaces. Generally the average incremental performance of \FFIL improves with more classes being introduced per incremental step, whereas the performance of \AFIL more-or-less stays consistent. In \autoref{fig:fig_AFIL_vs_FFIL_SwAV} we specifically provide the comparison between \FFIL and \AFIL using SwAV feature space, and instead of average, we provide the incremental accuracy for each batch of each incremental setting. For \FFIL, the 10-class incremental step performs better than 5-class, 5-class incremental step performs better than 2-class, and so on. This is because more data is available while fitting the EVMs in larger step sizes. For \AFIL, there is marginal improvement in performance due to smaller incremental steps.                 

\autoref{tab:incremental_OOL_OOD} provides the comparison of OOL and OOD settings where \AFIL is used for both types of self-supervised features. As expected, the OOL setting's performance where the unknowns belong to the same distribution as that employed for self-supervised feature extractor training outperforms the two OOD settings. Additionally, since the OpenImages data (object-centric) is closer to the ImageNet data (object-centric), the performance in this OOD setting is better compared to the Places2 data (scene-centric), which has less resemblance to ImageNet. This trend stands true for all our subsequent open-world experiments. 

In \autoref{tab:incremental_ResNet50_ResNet18}, we provide the comparison when MOCOv2 is trained using a ResNet-18 instead of ResNet-50 in an OOL setting and subsequently \AFIL is employed. We also list the results when ResNet-50/ResNet-18 is learned in a supervised fashion using the data belonging to the batch zero of the incremental setting \ie first 50 classes. 

Our two layer perceptron is comprised of an input layer, one hidden layer and a classification layer. The hidden layer contains the number of nodes equal to half of the dimension of the underlying feature vector and the number of nodes in the classification layer adopt according to the incremental step; which are fixed in case of incremental setting and varying in case of open-world setting. This is due to the fact the number of classes being enrolled in open-world learning depend upon the classes being detected as unknowns. Specifically the hidden layer contains 1024 and 256 nodes for features learned through ResNet-50 (2048 feature vector) and ResNet-18 (512 feature vector) respectively. The two layer perceptron is trained for 300 epoach for each incremental phase; where we use a learning rate of 1e-2 for phase zero and 1e-3 for all subsequent phases.        

\begin{table}[h]
\capt{tab:diff_feat_incremental1}{\FFIL augmented with various self-supervised techniques}{Average incremental Top-1 accuracy for varying self-supervised techniques with the ImageNet-100 protocol. Second column shows the Top-1 classification accuracy on ImageNet-2012 validation set. For each incremental setting, we also list the best parameter combination in the for form of (dm,ct) for the underlying EVM model. The best parameter has been found through a grid search for each feature space and each incremental setting using part of the training data. The last four rows demonstrate the performance sensitivity to these parameters.}
\centering
\small
\setlength{\arrayrulewidth}{.1em}
\setlength\tabcolsep{4.5pt}
\begin{tabularx}{\linewidth}{P{2.2cm}|P{.8cm}|cccc}
 \multirow{3}{*}{\textbf{\textit{Approach}}} & {\textbf{\textit{\% Top-1}}} & \multicolumn{4}{c}{\textbf{\textit{New Classes per Step}}}\\
  &  \textbf{\textit{Acc}}& \textit{1} & \textit{2} & \textit{5} & \textit{10}\\
  &  & \textit{n=50} & \textit{n=25} & \textit{n=10} & \textit{n=5}\\\cline{1-6}
 \multirow{2}{*}{\scriptsize MOCOv1 \cite{He2020Momentum}}                  & 60.6  & 47.68 & 53.41 & 58.14 & 62.15\\
 &  & \scriptsize(0.8,0.7) & \scriptsize(0.7,0.8) & \scriptsize(0.6,0.8) & \scriptsize(0.6,0.8)\\ \cline{1-6}
 \multirow{2}{*}{\scriptsize SimCLR 1x \cite{Chen2020A}}                    & 69.1  & 56.72 & 59.46 & 67.38 & 69.78\\
 &  & \scriptsize(0.8,0.8) & \scriptsize(0.8,0.8) & \scriptsize(0.7,0.8) & \scriptsize(0.7,0.8)\\ \cline{1-6}
 \multirow{2}{*}{\scriptsize MOCOv2 \cite{Chen2020Improved}}                & 71.1  & 64.02 & 67.60 & 72.71 & 74.50\\
 &  & \scriptsize(0.7,0.8) & \scriptsize(0.7,0.7) & \scriptsize(0.6,0.6) & \scriptsize(0.6,0.3)\\ \cline{1-6}
 \multirow{2}{*}{\scriptsize SeLa-v2 \cite{Asano2020Self-labelling}}        & 71.8  & 60.86 & 65.75 & \textbf{75.62} & 77.90\\
 &  & \scriptsize(0.8,0.8) & \scriptsize(0.8,0.8) & \scriptsize(0.6,0.8) & \scriptsize(0.6,0.8)\\ \cline{1-6}
 \multirow{2}{*}{\scriptsize DeepCluster-v2 \cite{Caron2020Unsupervised}}   & 75.2 & 64.32 & \textbf{70.01} & 74.61 & 76.39\\
 &  & \scriptsize(0.8,0.8) & \scriptsize(0.8,0.8) & \scriptsize(0.8,0.8) & \scriptsize(0.8,0.8)\\ \cline{1-6}
 \multirow{2}{*}{\scriptsize SwAV \cite{Caron2020Unsupervised}}         & \textbf{75.3}  & \textbf{64.50} & 68.28 & 75.59 & \textbf{78.09}\\
 &  & \scriptsize(0.8,0.8) & \scriptsize(0.8,0.8) & \scriptsize(0.7,0.8) & \scriptsize(0.7,0.8)\\
 \cdashline{1-6}  
 
 \multirow{2}{*}{\scriptsize SwAV \cite{Caron2020Unsupervised} par. var.}   & - & 63.20 & 67.59 & 75.51 & 77.86\\
 &  & \scriptsize(0.8,0.7) & \scriptsize(0.8,0.7) & \scriptsize(0.7,0.7) & \scriptsize(0.7,0.7)\\
 \multirow{2}{*}{\scriptsize SwAV \cite{Caron2020Unsupervised} par. var.}   & - & 65.72 & 69.57 & 76.14 & 78.54\\
 &  & \scriptsize(0.8,0.9) & \scriptsize(0.8,0.9) & \scriptsize(0.7,.9) & \scriptsize(0.7,0.9)\\
 \multirow{2}{*}{\scriptsize SwAV \cite{Caron2020Unsupervised} par. var.}   & - & 61.64 & 70.33 & 71.95 & 73.18\\
 &  & \scriptsize(0.7,0.8) & \scriptsize(0.7,0.8) & \scriptsize(0.6,0.8) & \scriptsize(0.6,0.8)\\
 \multirow{2}{*}{\scriptsize SwAV \cite{Caron2020Unsupervised} par. var.}   & - & 57.99 & 59.73 & 72.04 & 74.28\\
 &  & \scriptsize(0.9,0.8) & \scriptsize(0.9,0.8) & \scriptsize(0.8,0.8) & \scriptsize(0.8,0.8)\\
\end{tabularx}
\end{table}

\begin{table}[h]
\capt{tab:diff_feat_incremental2}{\AFIL augmented with various self-supervised techniques}{Average incremental Top-1 accuracy for varying self-supervised techniques with the ImageNet-100 protocol. The upper bound for best performing self-supervised feature space, \ie, SwAV \cite{Caron2020Unsupervised} is listed in last row where instead of 20 exemplars, all the data for all classes is used to train the two-layer perceptron in each incremental stage. The performance of the supervised feature space is listed in first row where the underlying network (ResNet-50) is trained in a supervised manner using data belong to first 50 classes.}
\centering
\small
\setlength{\arrayrulewidth}{.1em}
\setlength\tabcolsep{7.0pt}
\begin{tabularx}{\linewidth}{P{2.7cm}|cccc}
 \multirow{2}{*}{\textbf{\textit{Approach}}}  & \multicolumn{4}{c}{\textbf{\textit{New Classes per Step}}}\\
  & \textit{1} & \textit{2} & \textit{5} & \textit{10}\\
  & \textit{n=50} & \textit{n=25} & \textit{n=10} & \textit{n=5}\\\hline
 Supervised-B0                                  & 64.14 & 63.52 & 63.17 & 63.54\\ \cline{1-5}
 MOCOv1 \cite{He2020Momentum}                   & 67.64 & 68.44 & 69.81 & 70.60\\
 SimCLR 1x \cite{Chen2020A}                     & 78.21 & 77.58 & 77.30 & 77.02\\
 MOCOv2 \cite{Chen2020Improved}                 & 80.12 & 80.74 & 81.81 & 82.45\\
 SeLa-v2 \cite{Asano2020Self-labelling}         & 83.32 & 83.21 & 83.39 & 83.21\\
 DeepCluster-v2 \cite{Caron2020Unsupervised}    & 84.10 & 83.49 & 83.27 & 83.13 \\
 SwAV \cite{Caron2020Unsupervised}              & \textbf{84.86} & \textbf{84.55} & \textbf{84.30} & \textbf{84.28}\\ \cline{1-5}
 SwAV \cite{Caron2020Unsupervised} - upper bound& 89.87 & 89.77 & 89.75 & 89.69\\
\end{tabularx}
\end{table}

\begin{table}[h]
\capt{tab:incremental_OOL_OOD}{\AFIL trained on MOCOv2 \cite{Chen2020Improved} features in OOL and OOD settings.}{Average incremental Top-1 accuracy for OOL and OOD setting on ImageNet-100 protocol. OOL outperforms both OOD settings.}
\centering
\small
\setlength{\arrayrulewidth}{.1em}
\setlength\tabcolsep{7.0pt}
\begin{tabularx}{\linewidth}{P{2.7cm}|cccc}
 \multirow{2}{*}{\textbf{\textit{Approach}}} & \multicolumn{4}{c}{\textbf{\textit{New Classes per Step}}}\\
  &  \textit{1} & \textit{2} & \textit{5} & \textit{10}\\
  &  \textit{n=50} & \textit{n=25} & \textit{n=10} & \textit{n=5}\\ \cline{1-5}%\hline
 OOL-ImageNet           & 80.12 & 80.74 & 81.81 & 82.45\\
 OOD-OpenImages         & 66.75 & 67.37 & 68.94 & 70.06\\
 OOD-Places             & 62.11 & 63.06 & 64.45 & 65.85\\
 
\end{tabularx}
\end{table}

\begin{table}[h]
\capt{tab:incremental_ResNet50_ResNet18}{\AFIL trained on MOCOv2 \cite{Chen2020Improved} features learned through ResNet-50 vs ResNet-18.}{Average incremental Top-1 accuracy with the ImageNet-100 protocol.  Understandably \AFIL on ResNet-50 based features outperforms ResNet-18 features. Results for supervised feature spaces are also listed for both networks.}
\centering
\small
\setlength{\arrayrulewidth}{.1em}
\setlength\tabcolsep{7.0pt}
\begin{tabularx}{\linewidth}{P{2.7cm}|cccc}
 \multirow{2}{*}{\textbf{\textit{Approach}}} & \multicolumn{4}{c}{\textbf{\textit{New Classes per Step}}}\\
  & \textit{1} & \textit{2} & \textit{5} & \textit{10}\\
  & \textit{n=50} & \textit{n=25} & \textit{n=10} & \textit{n=5}\\\cline{1-5}
 
 Supervised-B0 \scriptsize(R-50)    & 64.14 & 63.52 & 63.17 & 63.54\\
 MOCOv2 \scriptsize(R-50)           & 80.12 & 80.74 & 81.81 & 82.45\\ \cline{1-5}
 Supervised-B0 \scriptsize(R-18)    & 60.80 & 60.16 & 60.02 & 60.66\\
 MOCOv2 \scriptsize(R-18)           & 63.89 & 65.26 & 66.46 & 67.48\\
\end{tabularx}
\end{table}

\begin{figure*}[h]
\begin{center}
\includegraphics[width=1.0\linewidth]{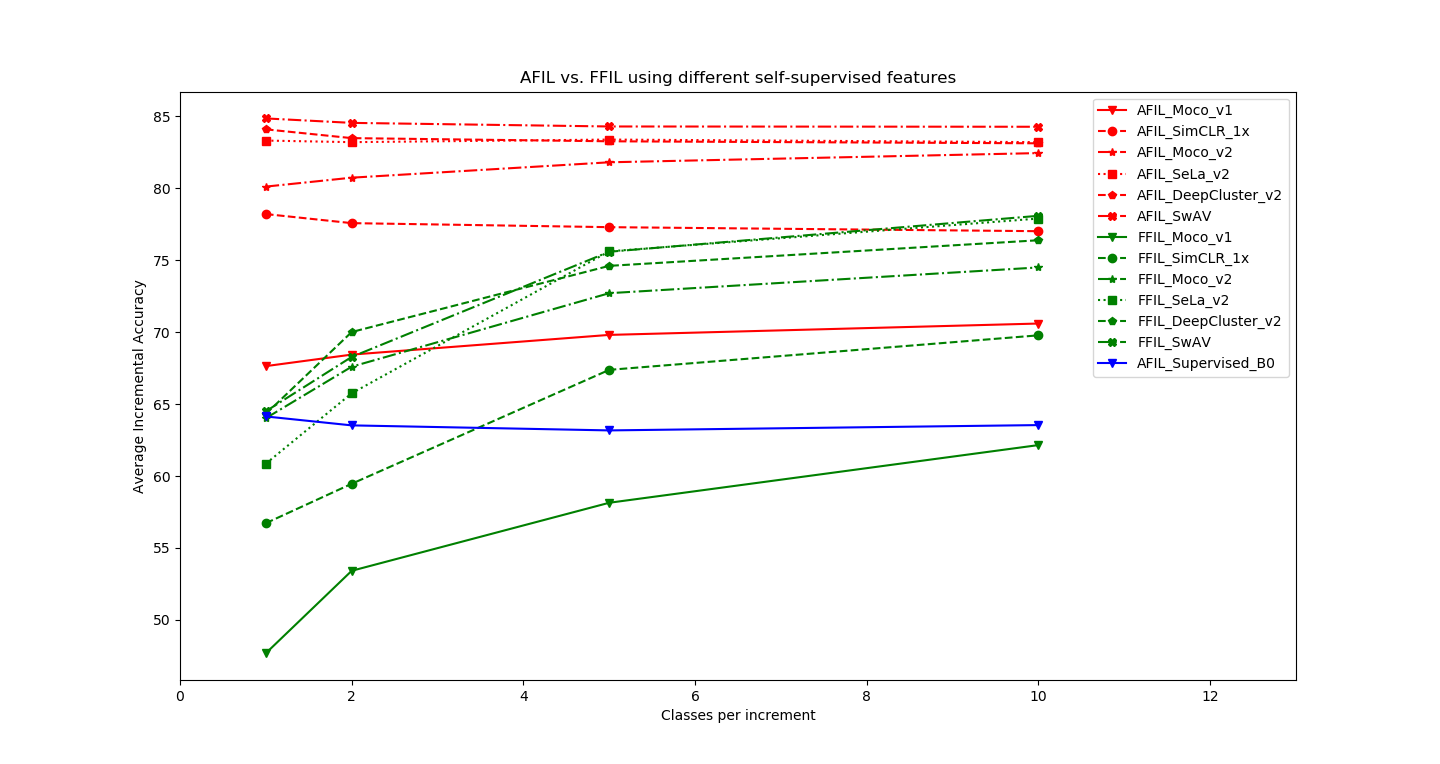}
\end{center}
\capt{fig:fig_AFIL_vs_FFIL}{AFIL vs. FFIL on all feature spaces}{Average incremental accuracy for \FFIL \& \AFIL using different feature spaces in different incremental step settings. SwAV outperforms others in both \FFIL and \AFIL, whereas MOCOv1 is the least performer.    
}
\end{figure*}

\begin{figure*}[h]
\begin{center}
\includegraphics[width=1.0\linewidth]{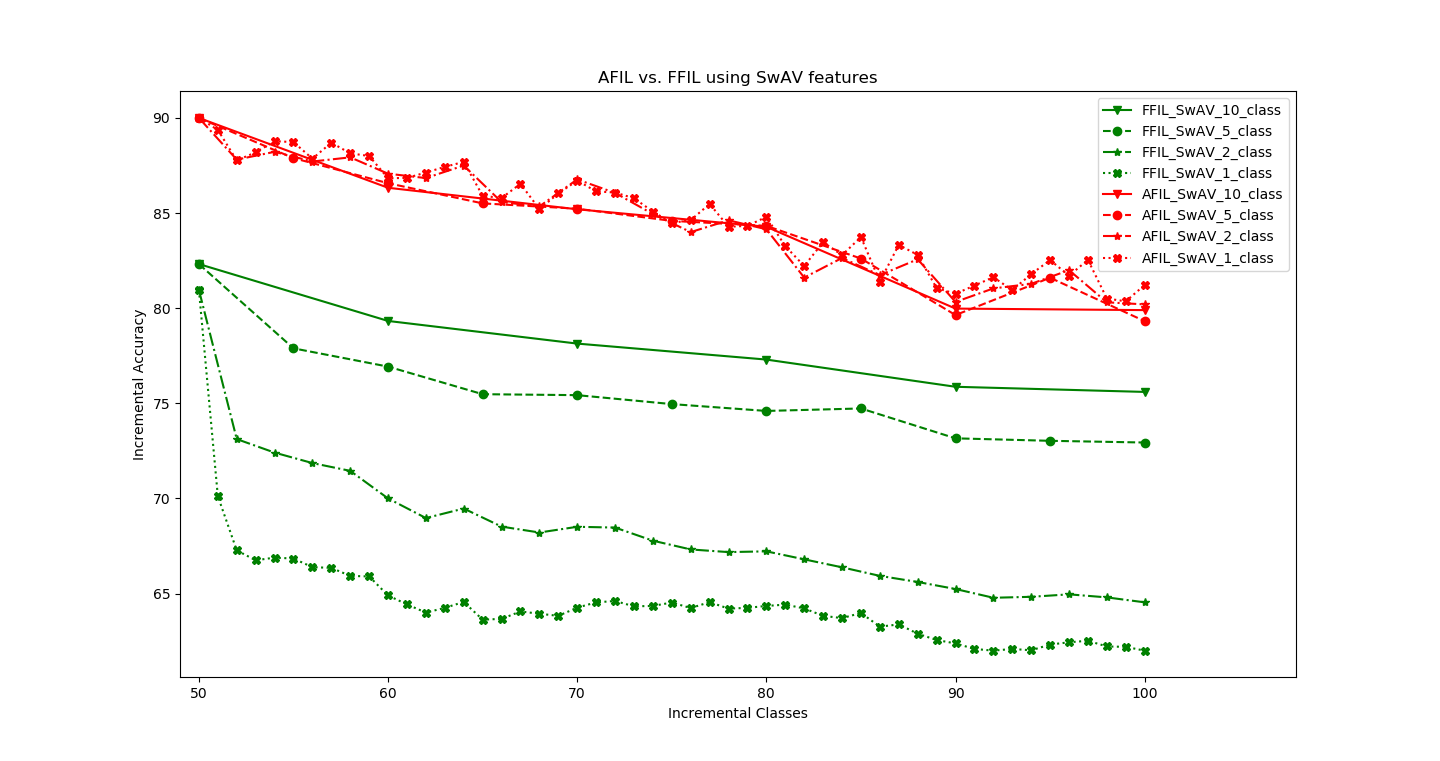}
\end{center}
\capt{fig:fig_AFIL_vs_FFIL_SwAV}{AFIL vs. FFIL on SwAV}{Incremental accuracy for \FFIL \& \AFIL using SwAV feature space. The performance after each incremental step is provided and generally declines after zeroth batch. For \FFIL the 10-class incremental step performs better than 5-class, 5-class incremental step performs better than 2-class and so on.        
}
\end{figure*}

\subsection*{Open-World Learning}
\autoref{tab:openworld_resnet50_OW-100} through \autoref{tab:openworld_resnet18_OW-500} provide the results for our open-world approaches \ie \FFOWL and \AFOWL. We have used MOCOv2 features instead of SwAV, regardless of SwAV outperforming all other feature spaces in incremental learning for our open-world experiments. This is due to the ease of training MOCOv2 models on custom datasets, but the same could be done for SWAV in the future.

\autoref{tab:openworld_resnet50_OW-100} and \autoref{tab:openworld_resnet50_OW-100_FFOWL} respectively provide the performance comparison for \AFOWL and \FFOWL on OW-100 protocol using introduced open-world metrics in OOL and both OOD settings \ie OpenImages and Places. The same information in graphical form is provided in \autoref{fig:fig_AFOWL_OW_100} and \autoref{fig:fig_FFOWL_OW_100}. Generally, for both \AFOWL and \FFOWL, the 10-class incremental setting outperforms the 5-class incremental setting. Consistently with incremental results, the OOL in both \AFOWL and \FFOWL cases outperforms both OOD settings irrespective of 10 or 5 class increments. Regardless of the underlying open-world approach, the performance gap of ImageNet-OpenImages is always smaller than the performance gap of ImageNet-Places as the OpenIamges dataset is semantically closer to ImageNet (object-centric) than that of Places (scene-centric). \autoref{fig:fig_AFOWL_vs_FFOWL_OW_100} provides a graphical comparison between \AFOWL and \FFOWL on OW-100 protocol in OOL setting. Consistent with incremental learning results, \AFOWL outperforms \FFOWL in both 5 and 10 class incremental settings.    

Next \autoref{tab:openworld_resnet50_OW-500} and \autoref{tab:openworld_resnet50_OW-500_FFOWL} respectively provide the performance comparison for \AFOWL and \FFOWL on OW-500 protocol, and same information in graphical form is depicted in \autoref{fig:fig_AFOWL_OW_500} and \autoref{fig:fig_FFOWL_OW_500}. Compared to OW-100 protocol, the performance on OW-500 declines for both \AFOWL and \FFOWL approaches regardless of the number of classes per increments and OOL/OOD setting. This is understandable as much more new classes are introduced incrementally in the case of OW-500 (450 incremental classes) than that of OW-100 (50 incremental classes), and the underlying models suffer more from catastrophic forgetting in the case of OW-500. \autoref{fig:fig_OW_100_vs_OW_500} provides a visual comparison of the performance of \AFOWL on OW-100 and OW-500 protocols in an OOL setting.

For underlying self-supervised network architecture variation, we have trained a ResNet-18 using MOCOv2 for OOL (ImageNet) and both OOD (OpenImages and Places) settings. \autoref{tab:openworld_resnet18_OW-100} \& \autoref{tab:openworld_resnet18_OW-500} provide the results for \AFOWL approach when a ResNet-18 instead of ResNet-50 is used to learn the underlying self-supervised feature space. Specifically \autoref{tab:openworld_resnet18_OW-100} and \autoref{tab:openworld_resnet18_OW-500} list the results for OW-100 and OW-500 protocols. As demonstrated in all ResNet-50 experiments, \AFOWL consistently outperforms \FFOWL, so \FFOWL results with ResNet-18 are not listed.

An acute reader might notice the absence of some entries in \autoref{tab:openworld_resnet50_OW-100} through \autoref{tab:openworld_resnet18_OW-500}. This is because as we compare OwCA and CwCA at a fixed UDA percentage, in some cases, it might happen that the approach did not achieve a required UDA percentage.  

\begin{table}[t!]
\capt{tab:openworld_resnet50_OW-100}{ResNet-50 -- \AFOWL OpenWorld Performance}{
Below we report the performance of our \AFOWL approach on the proposed OpenWorld-100 (OW-100) .
Based on the unlabeled data used during training of these networks, we report Out-Of-Label (OOL) space and Out-Of-Distribution (OOD).
The reported numbers are average accuracies across batches.
}
\centering
\small
\setlength{\arrayrulewidth}{0.1em}
\setlength\tabcolsep{2pt}
\begin{tabularx}{\linewidth}{M{1.8cm}|M{0.5cm}|cc!{\vrule width 0.05em}c|cc!{\vrule width 0.05em}c}
 \multirow{4}{*}{\textbf{\shortstack[lb]{Protocol\\Details}}} & \multirow{4}{*}{\rotatebox[origin=b]{90}{\textbf{\# Exemplars}}} & \multicolumn{6}{c}{\textbf{New classes per step $|U_n|$}}\\\cline{3-8}

 & & \multicolumn{3}{c|}{5} & \multicolumn{3}{c}{10}\\
 & & \multicolumn{3}{c|}{\textit{\scriptsize{\# Knowns/ \# Unknowns}}} & \multicolumn{3}{c}{\textit{\scriptsize{\# Knowns/ \# Unknowns}}}\\
 & & \multicolumn{3}{c|}{\textit{2500/1250}} & \multicolumn{3}{c}{\textit{2500/2500}}\\\cline{3-8}
 & & UDA & OwCA & CwCA & UDA & OwCA & CwCA\\\hline
 \multirow{2}{*}{OW-100 {\scriptsize (OOL)}} 
 & 20 & 30.01 & 73.20 & 78.47 & 30.01 & 67.92 & 78.05 \\
 & 20 & 40.02 & 71.42 & 76.57 & 40.02 & 66.40 & 76.31 \\
 & 20 & 50.02 & 68.32 & 73.24 & 50.02 & 64.27 & 73.84 \\
 & 20 & 60.02 & 64.02 & 68.63 & 60.02 & 61.16 & 70.26 \\
 & 20 & 70.03 & 57.81 & 61.98 & 70.03 & 56.08 & 64.42 \\
 & 20 & 80.03 & 48.62 & 52.11 & 80.03 & 47.68 & 54.75 \\
 & 20 & 90.04 & 35.00 & 37.52 & 90.04 & 33.65 & 38.69 \\
 & 20 & 95.20 & 25.96 & 27.80 & 95.04 & 23.14 & 26.65 \\\noalign{\hrule height 0.05em}
 \multirow{2}{*}{\shortstack[lb]{OW-100 {\scriptsize (OOD)}\\OpenImages}} 
 & 20 & 30.00 & 59.68 & 64.02 & 30.00 & 57.48 & 66.12 \\
 & 20 & 40.00 & 56.79 & 60.92 & 40.00 & 55.44 & 63.77 \\
 & 20 & 50.00 & 52.66 & 56.50 & 50.00 & 52.19 & 60.04 \\
 & 20 & 60.00 & 48.71 & 52.25 & 60.00 & 48.43 & 55.71 \\
 & 20 & 70.00 & 43.01 & 46.14 & 70.00 & 43.14 & 49.62 \\
 & 20 & 80.00 & 35.37 & 37.94 & 80.00 & 35.37 & 40.64 \\
 & 20 & 90.00 & 25.85 & 27.72 & 90.00 & 24.74 & 28.39 \\
 & 20 & 95.20 & 17.85 & 19.14 & 95.00 & 16.26 & 18.66 \\\noalign{\hrule height 0.05em}
 \multirow{2}{*}{\shortstack[lb]{OW-100 {\scriptsize (OOD)}\\Places}} 
 & 20 & 30.00 & 56.08 & 60.17 & 30.00 & 53.75 & 61.85 \\
 & 20 & 40.00 & 53.85 & 57.79 & 40.00 & 51.69 & 59.49 \\
 & 20 & 50.00 & 50.98 & 54.71 & 50.00 & 49.33 & 56.77 \\
 & 20 & 60.00 & 47.17 & 50.63 & 60.00 & 46.42 & 53.43 \\
 & 20 & 70.00 & 42.36 & 45.47 & 70.00 & 42.13 & 48.50 \\
 & 20 & 80.00 & 35.45 & 38.07 & 80.00 & 35.71 & 41.10 \\
 & 20 & 90.00 & 26.24 & 28.17 & 90.00 & 25.66 & 29.51 \\
 & 20 & 95.20 & 17.54 & 18.82 & 95.00 & 17.97 & 20.67 \\
\end{tabularx}
\end{table}

\begin{table}[t!]
\capt{tab:openworld_resnet50_OW-100_FFOWL}{ResNet-50 -- \FFOWL OpenWorld Performance}{
Below we report the performance of the our \FFOWL approach on the proposed OpenWorld-100 (OW-100) .
Based on the unlabeled data used during training of these networks we report Out-Of-Label (OOL) space and Out-Of-Distribution (OOD).
The reported numbers are average accuracies across batches.
}
\centering
\small
\setlength{\arrayrulewidth}{.1em}
\setlength\tabcolsep{2pt}
\begin{tabularx}{\linewidth}{M{1.8cm}|M{0.5cm}|cc!{\vrule width 0.05em}c|cc!{\vrule width 0.05em}c}
 \multirow{4}{*}{\textbf{\shortstack[lb]{Protocol\\Details}}} & \multirow{4}{*}{\rotatebox[origin=b]{90}{\textbf{\# Exemplars}}} & \multicolumn{6}{c}{\textbf{New classes per step $|U_n|$}}\\\cline{3-8}

 & & \multicolumn{3}{c|}{5} & \multicolumn{3}{c}{10}\\
 & & \multicolumn{3}{c|}{\textit{\scriptsize{\# Knowns/ \# Unknowns}}} & \multicolumn{3}{c}{\textit{\scriptsize{\# Knowns/ \# Unknowns}}}\\
 & & \multicolumn{3}{c|}{\textit{2500/1250}} & \multicolumn{3}{c}{\textit{2500/2500}}\\\cline{3-8}
 & & UDA & OwCA & CwCA & UDA & OwCA & CwCA\\\hline
 \multirow{2}{*}{OW-100 {\scriptsize (OOL)}} 
 & 20 & 30.00 & 67.29 & 72.14 & 30.00 & 63.02 & 72.43 \\
 & 20 & 40.00 & 66.51 & 71.32 & 40.00 & 62.14 & 71.43 \\
 & 20 & 50.00 & 65.13 & 69.84 & 50.00 & 60.83 & 69.94 \\
 & 20 & 60.00 & 62.57 & 67.11 & 60.00 & 58.76 & 67.59 \\
 & 20 & 70.00 & 59.39 & 63.71 & 70.00 & 55.81 & 64.22 \\
 & 20 & 80.00 & 54.18 & 58.15 & 80.00 & 51.15 & 58.91 \\
 & 20 & - & - & - & 90.00 & 41.57 & 47.96 \\
 & 20 & - & - & - & - & - & - \\\noalign{\hrule height 0.05em}
 \multirow{2}{*}{\shortstack[lb]{OW-100 {\scriptsize (OOD)}\\OpenImages}} 
 & 20 & 30.00 & 46.03 & 49.38 & 30.00 & 44.68 & 51.43 \\
 & 20 & 40.00 & 43.80 & 47.00 & 40.00 & 42.67 & 49.13 \\
 & 20 & 50.00 & 41.39 & 44.42 & 50.00 & 40.38 & 46.52 \\
 & 20 & 60.00 & 38.16 & 40.95 & 60.00 & 37.08 & 42.73 \\
 & 20 & 70.00 & 34.25 & 36.75 & 70.00 & 32.80 & 37.83 \\
 & 20 & 80.00 & 28.49 & 30.56 & 80.00 & 27.24 & 31.42 \\
 & 20 & 90.00 & 20.57 & 22.05 & 90.00 & 18.79 & 21.65 \\
 & 20 & 95.20 & 14.62 & 15.67 & 95.00 & 13.08 & 15.03 \\\noalign{\hrule height 0.05em}
 \multirow{2}{*}{\shortstack[lb]{OW-100 {\scriptsize (OOD)}\\Places}} 
 & 20 & 30.00 & 38.88 & 41.72 & 30.00 & 38.15 & 43.93 \\
 & 20 & 40.00 & 36.50 & 39.18 & 40.00 & 35.92 & 41.41 \\
 & 20 & 50.00 & 34.07 & 36.57 & 50.00 & 33.45 & 38.59 \\
 & 20 & 60.00 & 30.72 & 32.99 & 60.00 & 29.96 & 34.58 \\
 & 20 & 70.00 & 27.16 & 29.17 & 70.00 & 26.65 & 30.78 \\
 & 20 & 80.00 & 22.79 & 24.48 & 80.00 & 21.94 & 25.35 \\
 & 20 & 90.00 & 16.65 & 17.88 & 90.00 & 16.14 & 18.66 \\
 & 20 & 95.20 & 12.21 & 13.11 & 95.00 & 11.04 & 12.75 \\
 
\end{tabularx}
\end{table}

\begin{table}[t!]
\capt{tab:openworld_resnet50_OW-500}{ResNet-50 -- \AFOWL OpenWorld Performance }{
Below we report the performance of the our \AFOWL approach on the proposed OpenWorld-500 (OW-500) .
Based on the unlabeled data used during training of these networks, we report Out-Of-Label (OOL) space and Out-Of-Distribution (OOD).
The reported numbers are average accuracies across batches.
}
\centering
\small
\setlength{\arrayrulewidth}{.1em}
\setlength\tabcolsep{2pt}
\begin{tabularx}{\linewidth}{M{1.8cm}|M{0.5cm}|cc!{\vrule width 0.05em}c|cc!{\vrule width 0.05em}c}
 \multirow{4}{*}{\textbf{\shortstack[lb]{Protocol\\Details}}} & \multirow{4}{*}{\rotatebox[origin=b]{90}{\textbf{\# Exemplars}}} & \multicolumn{6}{c}{\textbf{New classes per step $|U_n|$}}\\\cline{3-8}

 & & \multicolumn{3}{c|}{5} & \multicolumn{3}{c}{10}\\
 & & \multicolumn{3}{c|}{\textit{\scriptsize{\# Knowns/ \# Unknowns}}} & \multicolumn{3}{c}{\textit{\scriptsize{\# Knowns/ \# Unknowns}}}\\
 & & \multicolumn{3}{c|}{\textit{2500/1250}} & \multicolumn{3}{c}{\textit{2500/2500}}\\\cline{3-8}
 & & UDA & OwCA & CwCA & UDA & OwCA & CwCA\\\hline
 OW-500 {\scriptsize (OOL)}
 & 20 & 30.00 & 59.69 & 61.38 & 30.00 & 58.44 & 61.75 \\
 & 20 & 40.00 & 57.50 & 59.13 & 40.00 & 56.31 & 59.52 \\
 & 20 & 50.00 & 54.40 & 55.95 & 50.00 & 53.40 & 56.46 \\
 & 20 & 60.00 & 50.47 & 51.92 & 60.00 & 49.55 & 52.41 \\
 & 20 & 70.00 & 45.38 & 46.69 & 70.00 & 44.23 & 46.80 \\
 & 20 & 80.00 & 38.87 & 39.99 & 80.00 & 37.66 & 39.85 \\
 & 20 & 90.00 & 29.55 & 30.40 & 90.00 & 27.30 & 28.89 \\
 & 20 & 95.21 & 22.18 & 22.81 & - & - & - \\\noalign{\hrule height 0.05em}
 \multirow{2}{*}{\shortstack[lb]{OW-500 {\scriptsize (OOD)}\\OpenImages}} 
 & 20 & 30.00 & 40.78 & 42.02 & 30.00 & 40.67 & 43.16 \\
 & 20 & 40.00 & 38.72 & 39.90 & 40.00 & 38.70 & 41.07 \\
 & 20 & 50.00 & 36.23 & 37.33 & 50.00 & 36.24 & 38.48 \\
 & 20 & 60.00 & 33.32 & 34.34 & 60.00 & 33.23 & 35.28 \\
 & 20 & 70.00 & 29.54 & 30.45 & 70.00 & 29.38 & 31.21 \\
 & 20 & 80.00 & 24.48 & 25.22 & 80.00 & 24.35 & 25.85 \\
 & 20 & 90.00 & 17.43 & 17.97 & 90.00 & 17.04 & 18.09 \\
 & 20 & 95.20 & 11.97 & 12.35 & 95.00 & 11.12 & 11.82 \\\noalign{\hrule height 0.05em}
 \multirow{2}{*}{\shortstack[lb]{OW-500 {\scriptsize (OOD)}\\Places}} 
 & 20 & 30.00 & 36.52 & 37.65 & 30.00 & 36.95 & 39.24 \\
 & 20 & 40.00 & 34.82 & 35.89 & 40.00 & 35.27 & 37.46 \\
 & 20 & 50.00 & 32.73 & 33.75 & 50.00 & 33.19 & 35.26 \\
 & 20 & 60.00 & 30.08 & 31.02 & 60.00 & 30.54 & 32.46 \\
 & 20 & 70.00 & 26.73 & 27.57 & 70.00 & 27.20 & 28.92 \\
 & 20 & 80.00 & 22.45 & 23.16 & 80.00 & 22.73 & 24.18 \\
 & 20 & 90.00 & 16.16 & 16.68 & 90.00 & 16.21 & 17.25 \\
 & 20 & 95.20 & 11.24 & 11.60 & 95.00 & 11.01 & 11.73 \\
\end{tabularx}
\end{table}

\begin{table}[t!]
\capt{tab:openworld_resnet50_OW-500_FFOWL}{ResNet-50 -- \FFOWL OpenWorld Performance}{
Below we report the performance of the our \FFOWL approach on the proposed OpenWorld-500 (OW-500) .
Based on the unlabeled data used during training of these networks we report Out-Of-Label (OOL) space and Out-Of-Distribution (OOD).
The reported numbers are average accuracies across batches.
}
\centering
\small
\setlength{\arrayrulewidth}{.1em}
\setlength\tabcolsep{2pt}
\begin{tabularx}{\linewidth}{M{1.8cm}|M{0.5cm}|cc!{\vrule width 0.05em}c|cc!{\vrule width 0.05em}c}
 \multirow{4}{*}{\textbf{\shortstack[lb]{Protocol\\Details}}} & \multirow{4}{*}{\rotatebox[origin=b]{90}{\textbf{\# Exemplars}}} & \multicolumn{6}{c}{\textbf{New classes per step $|U_n|$}}\\\cline{3-8}

 & & \multicolumn{3}{c|}{5} & \multicolumn{3}{c}{10}\\
 & & \multicolumn{3}{c|}{\textit{\scriptsize{\# Knowns/ \# Unknowns}}} & \multicolumn{3}{c}{\textit{\scriptsize{\# Knowns/ \# Unknowns}}}\\
 & & \multicolumn{3}{c|}{\textit{2500/1250}} & \multicolumn{3}{c}{\textit{2500/2500}}\\\cline{3-8}
 & & UDA & OwCA & CwCA & UDA & OwCA & CwCA\\\hline
 \multirow{2}{*}{OW-500 {\scriptsize (OOL)}} 
 & 20 & 30.00 & 55.79 & 57.36 & 30.00 & 53.69 & 56.74 \\
 & 20 & 40.00 & 53.99 & 55.52 & 40.00 & 51.89 & 54.87 \\
 & 20 & 50.00 & 51.63 & 53.11 & 50.00 & 49.45 & 52.31 \\
 & 20 & 60.00 & 48.37 & 49.77 & 60.00 & 46.26 & 48.98 \\
 & 20 & 70.00 & 44.28 & 45.58 & 70.00 & 42.19 & 44.70 \\
 & 20 & 80.00 & 37.96 & 39.10 & 80.00 & 36.45 & 38.67 \\
 & 20 & - & - & - & - & - & - \\
 & 20 & - & - & - & - & - & - \\\noalign{\hrule height 0.05em}
 \multirow{2}{*}{\shortstack[lb]{OW-500 {\scriptsize (OOD)}\\OpenImages}} 
 & 20 & 30.00 & 33.34 & 34.32 & 30.00 & 32.80 & 34.76 \\
 & 20 & 40.00 & 31.17 & 32.10 & 40.00 & 30.49 & 32.33 \\
 & 20 & 50.00 & 28.68 & 29.54 & 50.00 & 27.93 & 29.64 \\
 & 20 & 60.00 & 25.70 & 26.48 & 60.00 & 24.85 & 26.39 \\
 & 20 & 70.00 & 22.27 & 22.95 & 70.00 & 21.36 & 22.70 \\
 & 20 & 80.00 & 18.17 & 18.73 & 80.00 & 17.27 & 18.37 \\
 & 20 & 90.00 & 12.64 & 13.04 & 90.00 & 11.81 & 12.56 \\
 & 20 & - & - & - & 95.00 & 7.98 & 8.49 \\\noalign{\hrule height 0.05em}
 \multirow{2}{*}{\shortstack[lb]{OW-500 {\scriptsize (OOD)}\\Places}} 
 & 20 & 30.00 & 28.79 & 29.63 & 30.00 & 28.30 & 29.98 \\
 & 20 & 40.00 & 26.72 & 27.51 & 40.00 & 26.09 & 27.66 \\
 & 20 & 50.00 & 24.41 & 25.13 & 50.00 & 23.67 & 25.10 \\
 & 20 & 60.00 & 21.72 & 22.37 & 60.00 & 20.99 & 22.28 \\
 & 20 & 70.00 & 18.77 & 19.33 & 70.00 & 17.84 & 18.95 \\
 & 20 & 80.00 & 15.23 & 15.69 & 80.00 & 14.27 & 15.17 \\
 & 20 & 90.00 & 10.72 & 11.05 & 90.00 & 10.04 & 10.68 \\
 & 20 & 95.20 & 7.63 & 7.87 & 95.00 & 6.97 & 7.42 \\
 
\end{tabularx}
\end{table}

\begin{figure*}[h!]
\begin{center}
\includegraphics[width=0.95\linewidth]{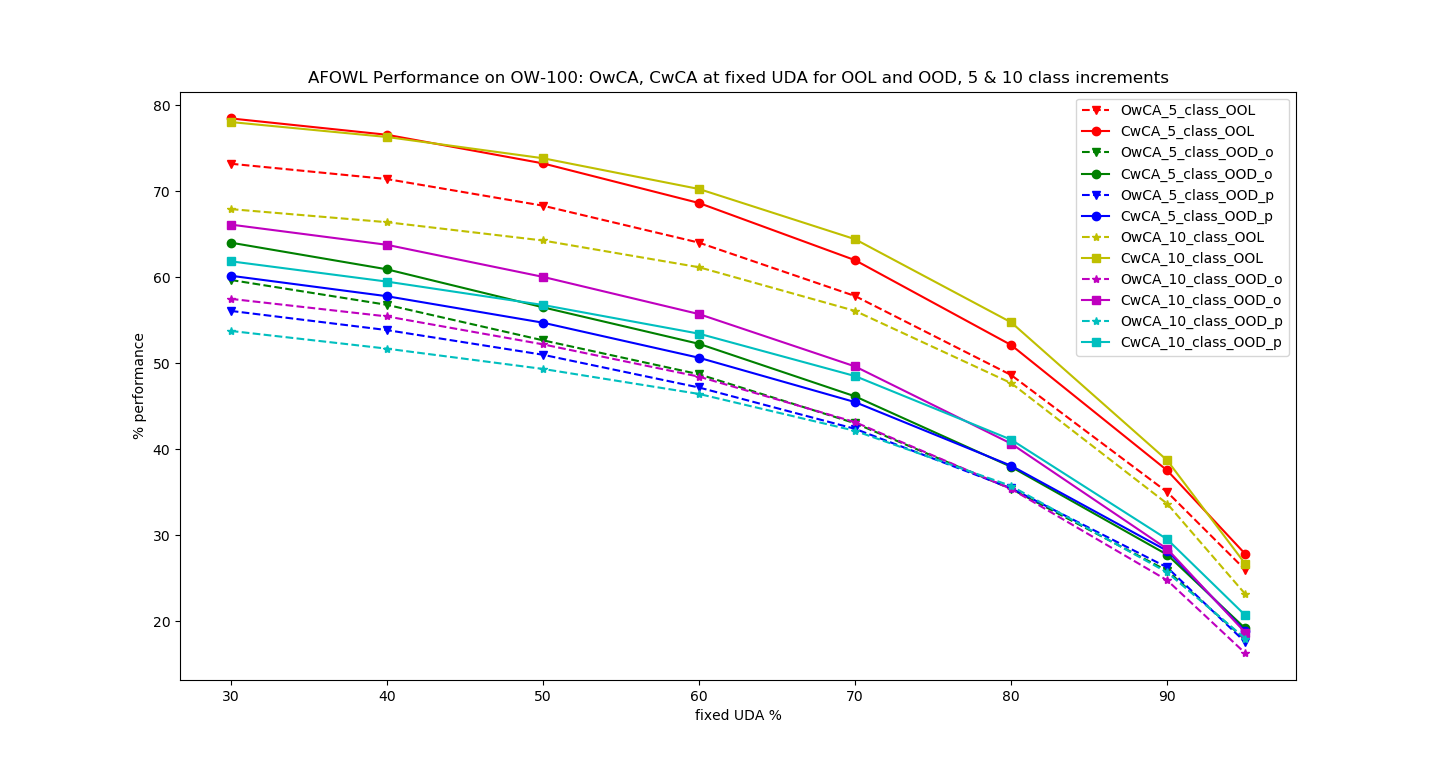}
\end{center}
\capt{fig:fig_AFOWL_OW_100}{AFOWL performance on OW-100}{OwCA and CwCA are reported for each fixed UDA percentage. Generally performance of 10-class increments is always greater than 5-class increments. The OOL setting regardless of number of incremental classes outperforms both OOD settings. The performance gap between ImageNet-OpenImages is smaller than the performance gap between ImageNet-Places. 
}
\end{figure*}

\begin{figure*}[h!]
\begin{center}
\includegraphics[width=0.95\linewidth]{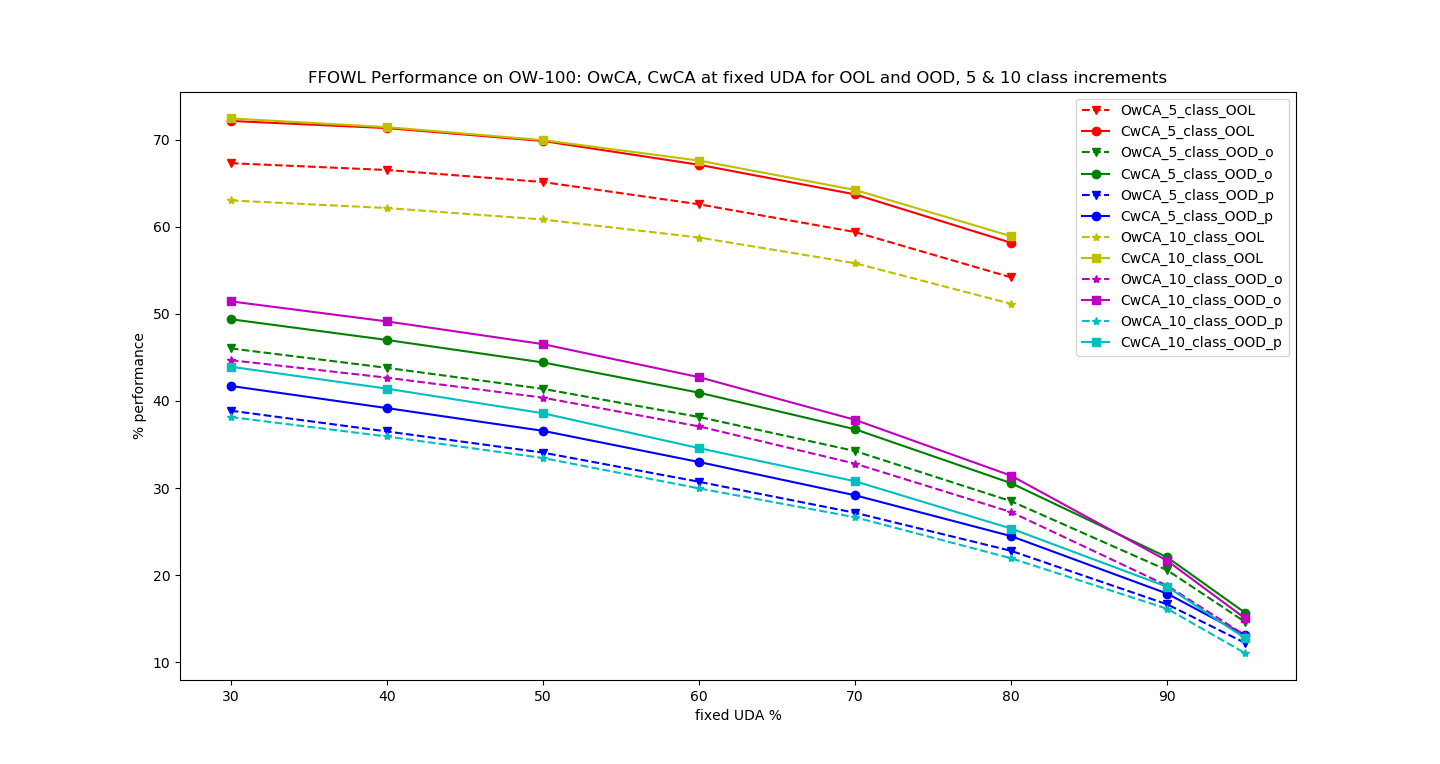}
\end{center}
\capt{fig:fig_FFOWL_OW_100}{FFOWL performance on OW-100}{OwCA and CwCA are reported for each fixed UDA percentage. Generally performance of 10-class increments is always greater than 5-class increments. The OOL setting regardless of number of incremental classes outperforms both OOD settings. The performance gap between ImageNet-OpenImages is smaller than the performance gap between ImageNet-Places.   
}
\end{figure*}

\begin{figure*}[h!]
\begin{center}
\includegraphics[width=0.95\linewidth]{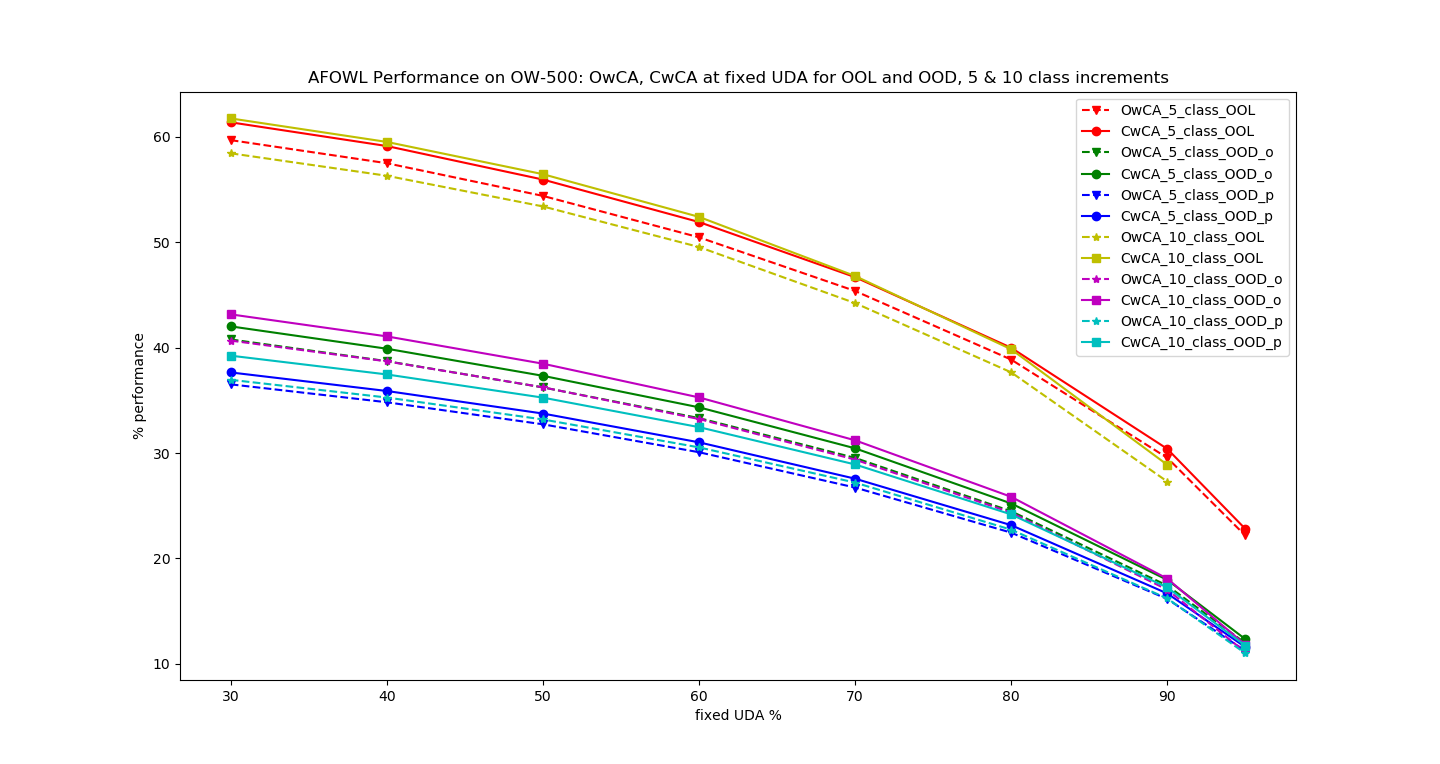}
\end{center}
\capt{fig:fig_AFOWL_OW_500}{AFOWL performance on OW-500}{OwCA and CwCA are reported for each fixed UDA percentage. Generally performance of 10-class increments is always greater than 5-class increments. The OOL setting regardless of number of incremental classes outperforms both OOD settings. The performance gap between ImageNet-OpenImages is smaller than the performance gap between ImageNet-Places.
}
\end{figure*}

\begin{figure*}[h!]
\begin{center}
\includegraphics[width=0.95\linewidth]{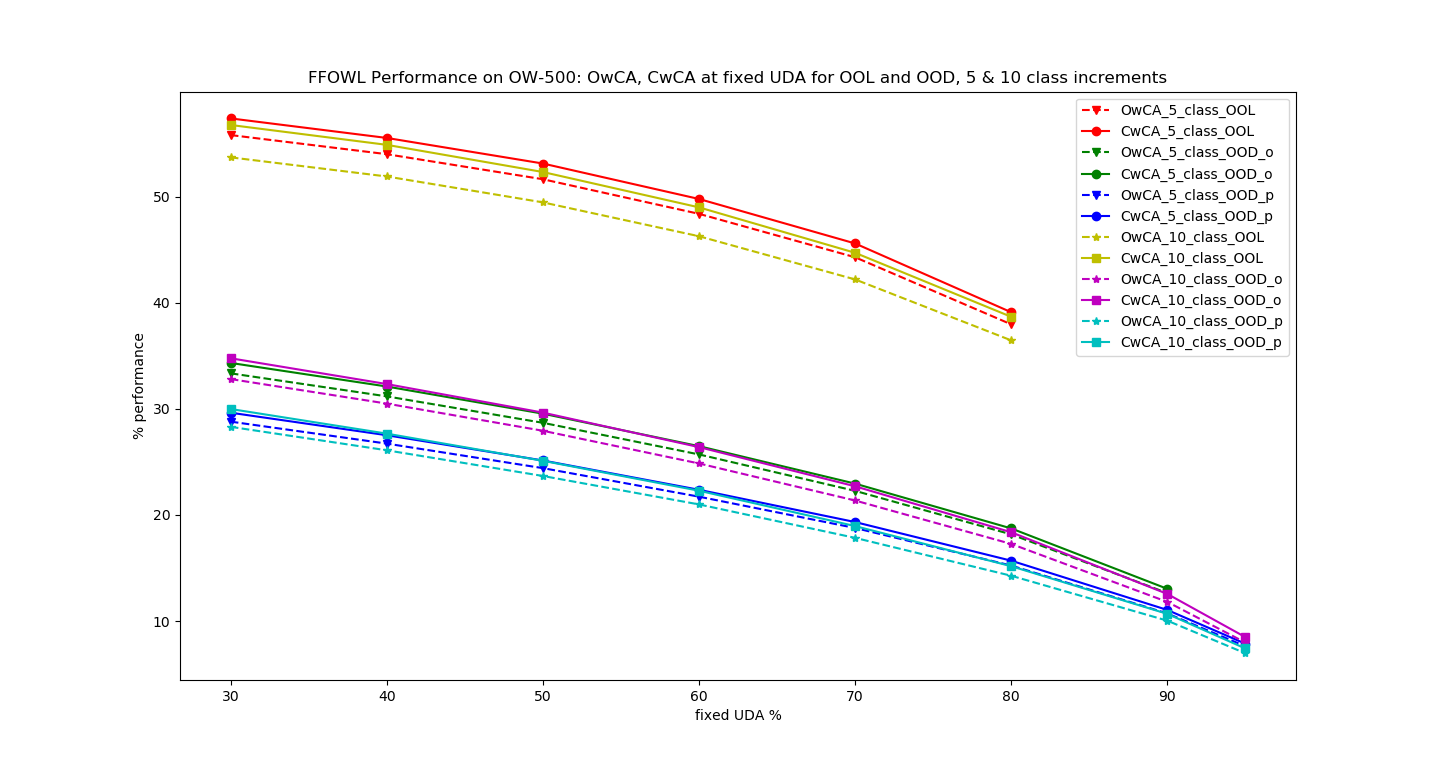}
\end{center}
\capt{fig:fig_FFOWL_OW_500}{FFOWL performance on OW-500}{OwCA and CwCA are reported for each fixed UDA percentage. Generally performance of 10-class increments is always greater than 5-class increments. The OOL setting regardless of number of incremental classes outperforms both OOD settings. The performance gap between ImageNet-OpenImages is smaller than the performance gap between ImageNet-Places.
}
\end{figure*}

\begin{figure*}[h!]
\begin{center}
\includegraphics[width=0.95\linewidth]{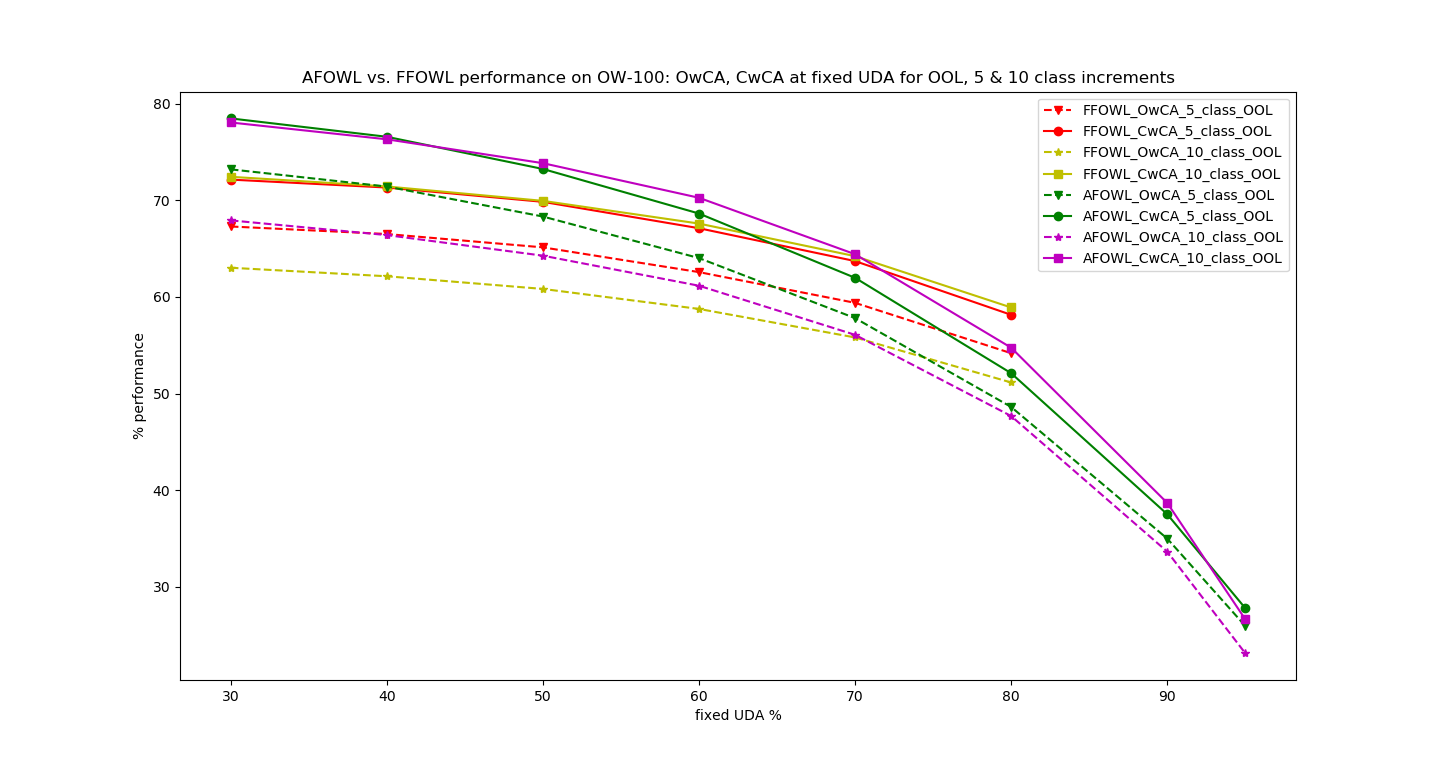}
\end{center}
\capt{fig:fig_AFOWL_vs_FFOWL_OW_100}{AFOWL vs. FFOWL on OW-100}{Performance comparison between \AFOWL and \FFOWL on OW-100 protocol using OOL setting and both variants of class increments. Generally \AFOWL outperforms \FFOWL in both incremental settings. 
}
\end{figure*}

\begin{figure*}[h!]
\begin{center}
\includegraphics[width=0.95\linewidth]{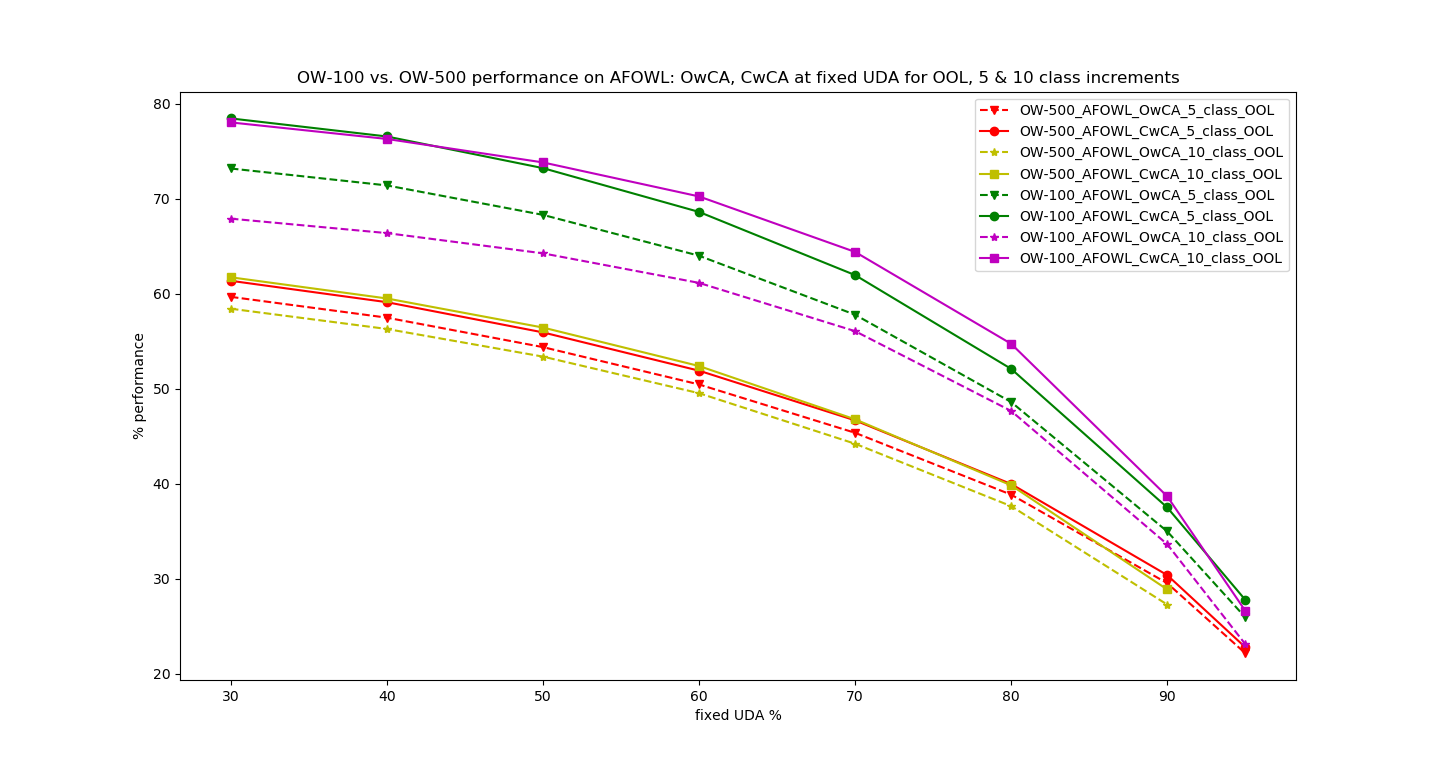}
\end{center}
\capt{fig:fig_OW_100_vs_OW_500}{OW-100 vs. OW-500 using AFOWL}{Performance comparison between OW-100 and OW-500 protocol using \AFOWL approach in OOL setting and both variants of class increments. Understandably performance of OW-500 lower than OW-100 due to large number of increments and consequently more catastrophic forgetting.  
}
\end{figure*}

\begin{table}[t!]
\capt{tab:openworld_resnet18_OW-100}{ResNet-18 -- \AFOWL OpenWorld Performance}{
Below we report the performance of the our \AFOWL approach on the proposed OpenWorld-100 (OW-100) .
Based on the unlabeled data used during these networks' training, we report Out-Of-Label space (OOL) and Out-Of-Distribution (OOD).
The reported numbers are average accuracies across batches.
}
\centering
\small
\setlength{\arrayrulewidth}{.1em}
\setlength\tabcolsep{2pt}
\begin{tabularx}{\linewidth}{M{1.8cm}|M{0.5cm}|cc!{\vrule width 0.05em}c|cc!{\vrule width 0.05em}c}
 \multirow{4}{*}{\textbf{\shortstack[lb]{Protocol\\Details}}} & \multirow{4}{*}{\rotatebox[origin=b]{90}{\textbf{\# Exemplars}}} & \multicolumn{6}{c}{\textbf{New classes per step $|U_n|$}}\\\cline{3-8}

 & & \multicolumn{3}{c|}{5} & \multicolumn{3}{c}{10}\\
 & & \multicolumn{3}{c|}{\textit{\scriptsize{\# Knowns/ \# Unknowns}}} & \multicolumn{3}{c}{\textit{\scriptsize{\# Knowns/ \# Unknowns}}}\\
 & & \multicolumn{3}{c|}{\textit{2500/1250}} & \multicolumn{3}{c}{\textit{2500/2500}}\\\cline{3-8}
 & & UDA & OwCA & CwCA & UDA & OwCA & CwCA\\\hline
 \multirow{2}{*}{OW-100 {\scriptsize (OOL)}} 
 & 20 & 30.00 & 55.17 & 59.20 & 30.00 & 53.03 & 61.03 \\
 & 20 & 40.00 & 52.52 & 56.35 & 40.00 & 50.85 & 58.53 \\
 & 20 & 50.00 & 49.26 & 52.86 & 50.00 & 48.21 & 55.50 \\
 & 20 & 60.00 & 45.00 & 48.29 & 60.00 & 44.29 & 50.98 \\
 & 20 & 70.00 & 39.75 & 42.67 & 70.00 & 39.00 & 44.90 \\
 & 20 & 80.00 & 31.20 & 33.50 & 80.00 & 31.44 & 36.20 \\
 & 20 & 90.00 & 20.57 & 22.10 & 90.00 & 19.76 & 22.80 \\
 & 20 & 95.20 & 14.05 & 15.09 & 95.00 & 13.07 & 15.11 \\\noalign{\hrule height 0.05em}
 \multirow{2}{*}{\shortstack[lb]{OW-100 {\scriptsize (OOD)}\\OpenImages}} 
 & 20 & 30.00 & 47.64 & 51.13 & 30.00 & 45.20 & 52.05 \\
 & 20 & 40.00 & 45.03 & 48.33 & 40.00 & 42.82 & 49.32 \\
 & 20 & 50.00 & 42.24 & 45.34 & 50.00 & 40.40 & 46.54 \\
 & 20 & 60.00 & 38.62 & 41.45 & 60.00 & 36.86 & 42.45 \\
 & 20 & 70.00 & 34.23 & 36.74 & 70.00 & 32.61 & 37.55 \\
 & 20 & 80.00 & 27.60 & 29.63 & 80.00 & 26.18 & 30.14 \\
 & 20 & 90.00 & 19.38 & 20.80 & 90.00 & 16.86 & 19.43 \\
 & 20 & 95.20 & 12.84 & 13.79 & 95.00 & 11.08 & 12.75 \\\noalign{\hrule height 0.05em}
 \multirow{2}{*}{\shortstack[lb]{OW-100 {\scriptsize (OOD)}\\Places}} 
 & 20 & 30.01 & 41.81 & 44.87 & 30.00 & 39.59 & 45.60 \\
 & 20 & 40.02 & 39.73 & 42.64 & 40.00 & 37.71 & 43.44 \\
 & 20 & 50.02 & 37.31 & 40.05 & 50.00 & 35.04 & 40.38 \\
 & 20 & 60.02 & 34.48 & 37.01 & 60.00 & 32.04 & 36.92 \\
 & 20 & 70.03 & 30.40 & 32.64 & 70.00 & 28.35 & 32.67 \\
 & 20 & 80.03 & 24.95 & 26.79 & 80.00 & 23.25 & 26.81 \\
 & 20 & 90.04 & 17.18 & 18.45 & 90.00 & 16.28 & 18.78 \\
 & 20 & 95.20 & 12.30 & 13.22 & 95.00 & 11.28 & 13.02 \\
\end{tabularx}
\end{table}

\begin{table}[t!]
\capt{tab:openworld_resnet18_OW-500}{ResNet-18 -- \AFOWL OpenWorld Performance}{
Below we report the performance of the our \AFOWL approach on the proposed OpenWorld-500 (OW-500) .
Based on the unlabeled data used during training of these networks, we report Out-Of-Label space (OOL) and Out-Of-Distribution (OOD).
The reported numbers are average accuracies across batches.
}
\centering
\small
\setlength{\arrayrulewidth}{.1em}
\setlength\tabcolsep{2pt}
\begin{tabularx}{\linewidth}{M{1.8cm}|M{0.5cm}|cc!{\vrule width 0.05em}c|cc!{\vrule width 0.05em}c}
 \multirow{4}{*}{\textbf{\shortstack[lb]{Protocol\\Details}}} & \multirow{4}{*}{\rotatebox[origin=b]{90}{\textbf{\# Exemplars}}} & \multicolumn{6}{c}{\textbf{New classes per step $|U_n|$}}\\\cline{3-8}

 & & \multicolumn{3}{c|}{5} & \multicolumn{3}{c}{10}\\
 & & \multicolumn{3}{c|}{\textit{\scriptsize{\# Knowns/ \# Unknowns}}} & \multicolumn{3}{c}{\textit{\scriptsize{\# Knowns/ \# Unknowns}}}\\
 & & \multicolumn{3}{c|}{\textit{2500/1250}} & \multicolumn{3}{c}{\textit{2500/2500}}\\\cline{3-8}
 & & UDA & OwCA & CwCA & UDA & OwCA & CwCA\\\hline
 OW-500 {\scriptsize (OOL)}
 & 20 & 30.00 & 37.73 & 38.87 & 30.00 & 37.60 & 39.89 \\
 & 20 & 40.00 & 35.68 & 36.76 & 40.00 & 35.66 & 37.84 \\
 & 20 & 50.00 & 33.13 & 34.14 & 50.00 & 33.17 & 35.21 \\
 & 20 & 60.00 & 29.95 & 30.87 & 60.00 & 30.08 & 31.94 \\
 & 20 & 70.00 & 26.10 & 26.91 & 70.00 & 26.13 & 27.75 \\
 & 20 & 80.00 & 21.22 & 21.87 & 80.00 & 20.99 & 22.30 \\
 & 20 & 90.00 & 14.72 & 15.17 & 90.00 & 14.17 & 15.04 \\
 & 20 & 95.20 & 10.23 & 10.55 & 95.00 & 9.70 & 10.30 \\\noalign{\hrule height 0.05em}
 \multirow{2}{*}{\shortstack[lb]{OW-500 {\scriptsize (OOD)}\\OpenImages}} 
 & 20 & 30.00 & 29.64 & 30.57 & 30.00 & 29.51 & 31.38 \\
 & 20 & 40.00 & 27.94 & 28.83 & 40.00 & 27.87 & 29.65 \\
 & 20 & 50.00 & 25.92 & 26.74 & 50.00 & 25.79 & 27.45 \\
 & 20 & 60.00 & 23.54 & 24.30 & 60.00 & 23.36 & 24.87 \\
 & 20 & 70.00 & 20.61 & 21.28 & 70.00 & 20.16 & 21.48 \\
 & 20 & 80.00 & 16.85 & 17.40 & 80.00 & 16.11 & 17.18 \\
 & 20 & 90.00 & 11.72 & 12.11 & 90.00 & 10.78 & 11.50 \\
 & 20 & 95.20 & 7.82 & 8.08 & 95.00 & 7.21 & 7.69 \\\noalign{\hrule height 0.05em}
 \multirow{2}{*}{\shortstack[lb]{OW-500 {\scriptsize (OOD)}\\Places}} 
 & 20 & 30.00 & 25.92 & 26.74 & 30.00 & 25.93 & 27.57 \\
 & 20 & 40.00 & 24.43 & 25.20 & 40.00 & 24.43 & 25.98 \\
 & 20 & 50.00 & 22.58 & 23.30 & 50.00 & 22.63 & 24.07 \\
 & 20 & 60.00 & 20.45 & 21.10 & 60.00 & 20.45 & 21.76 \\
 & 20 & 70.00 & 17.83 & 18.40 & 70.00 & 17.78 & 18.93 \\
 & 20 & 80.00 & 14.63 & 15.10 & 80.00 & 14.50 & 15.45 \\
 & 20 & 90.00 & 10.28 & 10.62 & 90.00 & 9.93 & 10.59 \\
 & 20 & 95.20 & 7.18 & 7.42 & 95.00 & 6.77 & 7.23 \\
\end{tabularx}
\end{table}

\end{document}